\documentclass{article}

\usepackage{amsmath,amsfonts,bm}

\def\eqref#1{equation~\ref{#1}}

\def\1{\bm{1}}

\def\vx{{\bm{x}}}

\def\mP{{\bm{P}}}

\DeclareMathAlphabet{\mathsfit}{\encodingdefault}{\sfdefault}{m}{sl}
\SetMathAlphabet{\mathsfit}{bold}{\encodingdefault}{\sfdefault}{bx}{n}

\usepackage{microtype}
\usepackage{graphicx}
\usepackage{subfigure}
\usepackage{booktabs} %
\usepackage{caption}
\usepackage{wrapfig}
\usepackage{multirow}
\usepackage{hyperref}
\usepackage{url}

\usepackage{hyperref}

\usepackage{makecell}
\usepackage{colortbl}
\definecolor{mygray}{gray}{.95}
\definecolor{mypink}{rgb}{.99,.91,.95}
\definecolor{mycyan}{cmyk}{.3,0,0,0}

\usepackage[accepted]{icml2023}
\usepackage{amsmath}
\usepackage{amssymb}
\usepackage{mathtools}
\usepackage{amsthm}

\usepackage{makecell}
\usepackage{colortbl}
\definecolor{mygray}{gray}{.95}
\definecolor{mypink}{rgb}{.99,.91,.95}
\definecolor{mycyan}{cmyk}{.3,0,0,0}

\newcommand{\dwj}[1]{{\color{orange}}}

\newcommand{\wj}[1]{{\color{black}#1}}

\newcommand{\ym}[1]{{\color{black}#1}}

\usepackage[capitalize,noabbrev]{cleveref}

\theoremstyle{plain}

\theoremstyle{definition}

\theoremstyle{remark}

\usepackage[textsize=tiny]{todonotes}

\icmltitlerunning{Characterizing Prediction Matrix for Unsupervised Accuracy Estimation}

\begin{document}

\twocolumn[
\icmltitle{Confidence and Dispersity Speak: Characterizing Prediction Matrix for Unsupervised Accuracy Estimation}

\icmlsetsymbol{equal}{*}

\begin{icmlauthorlist}
\icmlauthor{Weijian Deng}{anu}
\icmlauthor{Yumin Suh}{nec}
\icmlauthor{Stephen Gould}{anu}
\icmlauthor{Liang Zheng}{anu}
\end{icmlauthorlist}
\icmlaffiliation{nec}{NEC Laboratories America, Inc. (NEC Labs)}
\icmlaffiliation{anu}{The Australian National University}
\icmlcorrespondingauthor{Weijian Deng}{weijian.deng@anu.edu.au}
\icmlcorrespondingauthor{Yumin Suh}{yumin@nec-labs.com}
\icmlcorrespondingauthor{Stephen Gould}{gould.stephen@anu.edu.au}
\icmlcorrespondingauthor{Liang Zheng}{liang.zheng@anu.edu.au}

\vskip 0.3in
]

\printAffiliationsAndNotice{}  %

\begin{abstract}
This work aims to assess how well a model performs under distribution shifts without using labels. While recent methods study prediction confidence, this work reports prediction dispersity is another informative cue. Confidence reflects whether the \textit{individual} prediction is certain; dispersity indicates how the \textit{overall} predictions are distributed across all categories. Our key insight is that a well-performing model should give predictions with high confidence and high dispersity. That is, we need to consider both properties so as to make more accurate estimates. To this end, we use the nuclear norm that has been shown to be effective in characterizing both properties. Extensive experiments validate the effectiveness of nuclear norm for various models (\textit{e.g.}, ViT and ConvNeXt), different datasets (\textit{e.g.}, ImageNet and CUB-200), and diverse types of distribution shifts (\textit{e.g.}, style shift and reproduction shift). We show that the nuclear norm is more accurate and robust in accuracy estimation than existing methods. Furthermore, we validate the feasibility of other measurements (\emph{e.g.}, mutual information maximization) for characterizing dispersity and confidence. Lastly, we investigate the limitation of the nuclear norm, study its improved variant under severe class imbalance, and discuss potential directions.

\end{abstract}

\section{Introduction}

Model evaluation is critical in both machine learning research and practice. The standard evaluation protocol is to evaluate a model on a held-out test set that is 1) fully labeled and 2) drawn from the same distribution as the training set.
However, this way of evaluation is often infeasible for real-world deployment, where the test environments undergo distribution shifts and ground truths are not provided. %
In presence of a distribution shift, in-distribution accuracy may only be a weak predictor of model performance~\citep{deng2021labels,garg2022leveraging}.
Moreover, annotating data itself is a laborious task, let alone it is impractical to label every new test distribution.
Hence, a way to predict a classifier accuracy using unlabelled test data only has recently received much attention~\citep{chuang2020estimating,deng2021labels,guillory2021predicting,garg2022leveraging}.

In the task of accuracy estimation, existing methods typically derive model-based distribution statistics of test sets~\citep{deng2021labels,guillory2021predicting,Deng:ICML2021,deng2022strong,garg2022leveraging,baek2022agreement}.
Recent works develop methods based on prediction matrix on unlabeled data \citep{guillory2021predicting,garg2022leveraging}. They focus on the overall confidence of the prediction matrix.
{Confidence} refers to whether the model gives a confident prediction on individual test data.
It can be measured by entropy or maximum softmax probability.
\cite{guillory2021predicting} show that the average of maximum softmax scores on a test set is useful for accuracy estimation.
\cite{garg2022leveraging} predict accuracy as the fraction of test data with maximum softmax scores above a threshold.

In this work, we consider another property of prediction matrix: {dispersity}.
{It measures how spread out the predictions are across classes.}
When testing a source-trained classifier on a target (out-of-distribution) dataset, target features may exhibit degenerate structures due to the distribution shift, where many target features are distributed in a few clusters 
. 
As a result, their corresponding class predictions would also be degenerate rather than diverse: the classifier predicts test features into specific classes and few into others. %
There are existing works that encourage the cluster sizes in the target data to be balanced~\citep{shi2012information,liang2020we,yang2021exploiting,tang2020unsupervised}, thereby increasing the prediction dispersity.
In contrast, this work does not aim to improve cluster structures and instead studies the prediction dispersity to predict model accuracy on unlabeled test sets.

To illustrate that dispersity is useful for accuracy estimation, we report our empirical observation in Fig.~\ref{fig:dispersity}. 
{We compute the predicted dispersity score by measuring whether the frequency of the predicted class is uniform. Specifically, we use entropy to quantify the frequency distribution, with higher scores indicating that the overall predictions are well-balanced.}
We show that the dispersity score exhibits a very strong correlation (Spearman's rank correlation $\rho>0.950$) with classifier performance when testing on various test sets. This implies that when the classifier does not generalize well on the test set, it tends to give \textit{degenerate} predictions (\emph{i.e.}, low prediction dispersity), where the test samples are mainly assigned to some specific classes.

Based on the above observation, we propose to use nuclear norm, known to be effective in measuring both prediction dispersity and confidence \citep{cui2020towards,cui2021fast}, towards accurate estimation. 
{Other measurements can also be used, such as mutual information maximizing \citep{bridle1991unsupervised,krause2010discriminative,shi2012information}.}
Across various model architectures on a range of datasets, we show that the nuclear norm is more effective than state-of-the-art methods (\emph{e.g.}, ATC \citep{garg2022leveraging} and DoC \citep{guillory2021predicting}) in predicting OOD performance.
Using uncontrollable and severe synthetic corruptions, we show that nuclear norm is again superior. 
Finally, we demonstrate that the nuclear norm still makes reasonably accurate estimations for test sets with moderate imbalances of classes. We additionally discuss potential solutions under strong label shifts.

\section{Related Work}

\textbf{Unsupervised accuracy estimation} is proposed to evaluate a model on unlabeled datasets. Recent methods typically consider the characteristics of unlabeled test sets \citep{deng2021labels,guillory2021predicting,Deng:ICML2021,garg2022leveraging,baek2022agreement,yu2022predicting,chen2021mandoline,chen2021detecting}. For example, \cite{deng2021labels,yu2022predicting,chuang2020estimating} consider the distribution discrepancy for accuracy estimation.
\cite{chen2021mandoline} achieve more accurate estimation by using specified slicing functions in the importance weighting.
\cite{chuang2020estimating} learn a domain-invariant classifier on an unlabeled test set to estimate the target accuracy.
\cite{guillory2021predicting,garg2022leveraging}~propose to predict accuracy based on the softmax scores on unlabeled data. 
In addition, the agreement score of multiple models' predictions on test data is investigated in \citep{madani2004co,platanios2016estimating,platanios2017estimating,donmez2010unsupervised,chen2021detecting}.
This work also focuses on estimating a model's OOD accuracy on various datasets and proposes to achieve robust estimations by  considering both prediction confidence and dispersity.

\textbf{Predicting ID generalization gap.} To predict the performance gap between \textit{a certain pair} of the training-testing set, several works explore developing complexity measurements on trained models and training data \citep{eilertsen2020classifying,unterthiner2020predicting,arora2018stronger,corneanu2020computing,jiang2018predicting,neyshabur2017exploring,jiang2019fantastic,schiff2021predicting}.
For example, \cite{corneanu2020computing} predict the generalization gap by using persistent topology measures. 
\cite{jiang2018predicting} develop a measurement of layer-wise margin distributions for the generalization prediction.
\cite{neyshabur2017exploring} use the product of norms of the weights across multiple layers.
\cite{baldock2021deep} introduce a measure of example difficulty (\textit{i.e.}, prediction depth) to study the learning of deep models.
\cite{chuang2021measuring} develop margin-based generalization bounds with optimal transport.
The above works assume that the training and test sets are from the same distribution and they do not consider the characteristics of the test distribution. 
In comparison, we focus on predicting a model's accuracy on \textit{various} OOD datasets.

\textbf{Calibration} aims to make the probability obtained by the model reflect the true correctness likelihood \citep{guo2017calibration,minderer2021revisiting}. To achieve this, several methods have been developed to improve the calibration of their predictive uncertainty, both during training \citep{karandikar2021soft,krishnan2020improving} and after \citep{guo2017calibration,gupta2020calibration} training.
For a perfectly calibrated model, the average confidence over a distribution corresponds to its accuracy over this distribution. However, calibration methods seldom exhibit desired calibration performance under distribution shifts \citep{ovadia2019can,gong2021confidence}.
To estimate OOD accuracy, this work does not focus on calibrating confidence. Instead, we use the dispersity and confidence of prediction matrix to predict model performance on unlabeled data.

\section{Methodology}
\subsection{Problem Definition}
\textbf{Notations.} Consider a classification task with input space $\mathcal{X}\subseteq \mathbb{R}^d$ and label space $\mathcal{Y}=\{1,\dots,k\}$.
Let $p_{S}$ and $p_{T}$ denote source and target distributions over $\mathcal{X} \times \mathcal{Y}$, respectively.
\wj{Given a source training dataset $\mathcal{D}_\mathrm{train}^{S}$ drawn from $p_{S}$, we train a probabilistic predictor $f:\mathbb{R}^d \to \Delta_k$, where $\Delta_k$ denotes the $k-1$ dimensional unit simplex.}
We assume a held-out test set $\mathcal{D}_\mathrm{test}^{S}=\{(\vx^s_{i}, y^s_{i})\}_{i=1}^{n_s}$ contains $n_s$ data i.i.d sampled from $p_{S}$. 
When queried at source data $(\vx^s, y^s)$ of $\mathcal{D}_\mathrm{test}^{S}$, $f$ returns $\hat{y} \eqqcolon \arg\max_{j\in \mathcal{Y}}$ $f_{j}(\vx^s)$ as the predicted label and $\hat{p} \eqqcolon \max_{j\in \mathcal{Y}} f_{j} (\vx^s)$ as the associated softmax confidence score.
With label, we can easily compute the classification error on that data by $\mathcal{E}(f(\vx^s),y^s) \coloneqq \1_\mathrm{condition}(y^s\neq \hat{y})$. By calculating the errors on all data of $\mathcal{D}_\mathrm{test}^{S}$, we evaluate the accuracy $f$ on the source (in-distribution) $p_{S}$. 

\begin{figure*}[t]
    \begin{center}
    \includegraphics[width=1\linewidth]{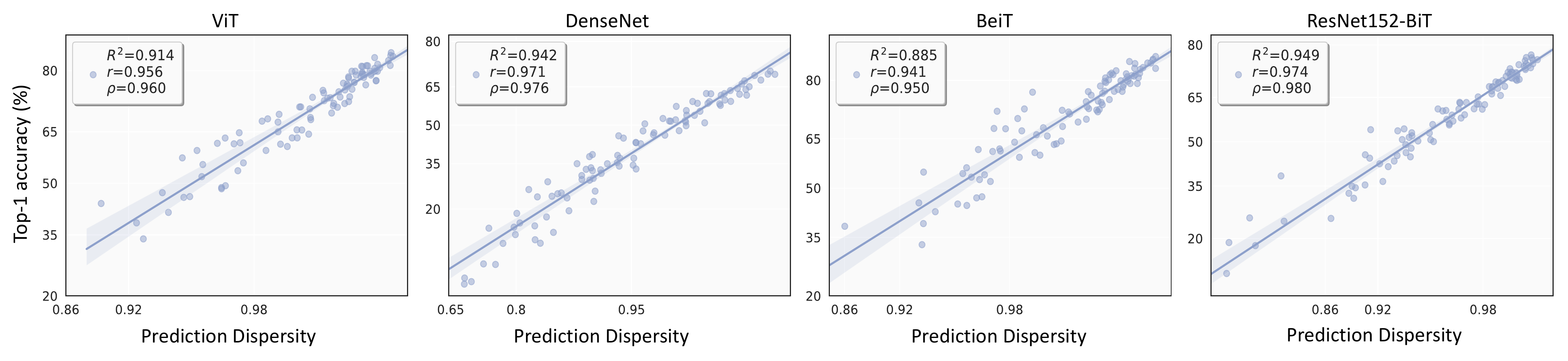}
    \vspace{-0.8cm}
    \caption{\textbf{Strong correlation between prediction dispersity and classifier accuracy.} 
    Each point corresponds to one test set of ImageNet-C.
    The straight lines are calculated by linear regression. 
    We study four ImageNet models (ViT, DenseNet, BeiT, and ResNet152-BiT).
    {We compute the predicted dispersity score by measuring how uniform the frequency of the predicted class is.}
    We observe that prediction dispersity exhibits a strong correlation (Spearman's rank correlation $\rho>0.950$) with classification accuracy for various test datasets.
    This indicates that if a classier gives class predictions with high dispersity, it likely achieves high accuracy, and not otherwise.
    \vspace{-0.5cm}
    }\label{fig:dispersity}
    \end{center}
\end{figure*}

\textbf{Unsupervised Accuracy Estimation.} Due to distribution shift ($p_S \neq p_T$), the accuracy on in-distribution $\mathcal{D}_\mathrm{test}^{S}$ is usually a weak estimate of how well $f$ performs on the target (out-of-distribution)~$p_T$.
This work aims to assess the generalization of $f$ on target (out-of-distribution) $p_{T}$ \emph{without access to labels}.
Concretely, given a source-trained $f$ and an unlabeled dataset $\mathcal{D}_\mathrm{u}^{T}=\{(\vx_i^{t})\}_{i=1}^{n_t}$ with $n_t$ samples  drawn i.i.d. from $p_{T}$, we aim to develop a quantity that strongly correlates with the accuracy of $f$ on $\mathcal{D}_\mathrm{u}^{T}$. Note that, the target distribution $p_{T}$ has the same $k$ classes as the source distribution $p_{S}$ in this work (known as the closed-set setting).
\wj{Unlike domain adaptation, which aims to adapt the model to the target data, unsupervised accuracy estimation focuses on predicting model accuracy on various unlabeled test sets}.

\subsection{Prediction Confidence and Dispersity}
Let $\mP\in\mathbb{R}^{n_t \times k}$ denote the prediction matrix of $f$ on $\mathcal{D}_\mathrm{u}^{T}$, \wj{and its each row $\mP_{i, :}$ is the softmax vector of $i$-th target data. The values of $\mP$ are in the interval [0, 1]}. 
Based on the predicted class of each softmax vector, we divide $\mP$ into $k$ class groups ($k$ is the number of classes). Then, we analyze the following two properties of $\mP$.
 
\textbf{Confidence} measures whether a softmax vector (each row of $\mP$) is certain. Common ways to measure confidence include entropy and maximum softmax score.
If the overall confidence of $\mP$ is high, then it implies that the classifier $f$ is certain on the given test set.
\textcolor{black}{Prediction confidence} has been reported to be useful in predicting classifier performance on various test sets \citep{guillory2021predicting,garg2022leveraging}.
For example, the overall confidence of $\mP$ measured by the average of maximum softmax score is predictive of classifier accuracy \citep{guillory2021predicting}. Other measures such as entropy \citep{garg2022leveraging} also give similar observations.

\textbf{Dispersity} measures whether the predicted classes are diverse and well-distributed.
\ym{High dispersity means that predictions on test samples are well-distributed among $k$ classes.}
When testing source-train classifier $f$ on a target dataset $\mathcal{D}_\mathrm{u}^{T}$, the target features may exhibit degenerate structures due to distribution shift. {A commonly seen pattern is that  many target features are distributed in few clusters.}
This likely leads to degenerate predictions: the classifier tends to predict test features into some particular classes (and \textcolor{black}{neglects other classes}). 
Recent methods \citep{tang2020unsupervised,liang2020we,yang2022attracting} report that regularizing prediction dispersity by encouraging cluster size to be balanced is beneficial when training domain adaptive models.
Here, we study whether prediction dispersity is useful for the problem of accuracy estimation, instead of adapting 
models to the target domain. 

\textcolor{black}{To verify the usefulness of dispersity in accuracy prediction}, we conduct preliminary correlation study using ImageNet-C in Fig.~\ref{fig:dispersity}. Here, the prediction dispersity score is simply computed by measuring whether the number of softmax vectors in each class is similar: we first calculate the histogram of the sizes of the predicted class and then use entropy to measure the degree of balance. 
We observe that prediction dispersity has a consistently strong correlation (rank correlation $\rho>0.950$) with model accuracy on various test sets (ImageNet-C).
This shows that when the classifier does not generalize well on test data, it tends to give \textit{degenerate} predictions (low prediction dispersity), where the test samples are mainly assigned to some specific categories.

\subsection{Characterizing Dispersity and Confidence with Nuclear Norm} \label{sec:nuclear_norm}
Based on the above observation, we aim to \ym{quantify} dispersity and confidence of prediction matrix $\mP$ for accuracy estimation.
For this purpose, we resort to the nuclear norm which is known to be effective in \ym{measuring both} prediction dispersity and confidence~\citep{cui2020towards,cui2021fast}. 

Nuclear norm $||\mP||_*$ is defined as the sum of singular values of $\mP$. It is the tightest convex envelope of rank function within the unit ball \citep{fazel2002matrix}.
\wj{A larger nuclear norm implies more classes are predicted and involved, indicating higher prediction dispersity.
In addition, the nuclear-norm $||\mP||_*$ is an upperbound of the Frobenius-norm that $||\mP||_F$ reflects prediction confidence \cite{cui2020towards}.} 
In Section \ref{supp:nuclear} of the appendix, we briefly introduce how nuclear norm reflects the prediction confidence and dispersity.
\ym{Since test sets can contain any number of data points, we normalize} \textcolor{black}{the nuclear norm of prediction matrix} by its upper bound derived from matrix size and obtain $\widehat{||\mP||_*} = ||\mP||_*/\sqrt{\min(n_t,k) \cdot n_t}$. 
We mainly use $\widehat{||\mP||_*}$ to measure the confidence and dispersity of prediction matrix $\mP$ in this work. 
\wj{In the experiment, we also show that another measure mutual information maximization \citep{bridle1991unsupervised,krause2010discriminative,shi2012information,yang2022attracting} is also feasible for the task of accuracy estimation.}

\section{Experiment}

\subsection{Experimental Setups} \label{sec:exp_setting}

\textbf{ImageNet-1K.} \underline{(i) Model.} We use $6$ representative neural networks provided by \citep{rw2019timm}. First, we include three vision transformers: ViT-Base-P16 (ViT) \citep{dosovitskiy2020image}, BEiT-Base-P16 (BEiT) \citep{liu2022convnet}, and Swin-Small-P16 (Swin) \citep{liu2021swin}. Second, we include three convolution neural networks: DenseNet-121 (DenseNet), ResNetv2-152-BiT-M (Res152-BiT) \citep{kolesnikov2020big}, ConvNeXt-Base \citep{liu2022convnet}. They are either trained or fine-tuned on ImageNet training set~\citep{deng2009imagenet}.
\underline{(ii) Synthetic Shift.} We use ImageNet-C benchmark \citep{hendrycks2019robustness} to study the synthetic distribution shift. ImageNet-C is controllable in terms of both type and intensity of corruption. It contains $95$ datasets that are generated by applying  $19$ types of corruptions (\textit{e.g.}, blur and contrast) to ImageNet validation set. Each type has $5$ intensity levels.
\underline{(iii) Real-world Shift.} We consider four natural shifts, including 1) dataset reproduction shift in ImageNet-V2-A/B/C~\citep{recht2019imagenet}, 2) sketch shift in  ImageNet-S(ketch)~\citep{wang2019learning}, 3) style shift in ImageNet-R(endition)~\citep{hendrycks2021many}, and 4) bias-controlled dataset shift in ObjectNet~\citep{barbu2019objectnet}. 
Note that, ImageNet-R and ObjectNet only share common $113$ and $200$ classes with ImageNet, respectively. Following \citep{hendrycks2021many}, we sub-select the model logits for the common classes of both test sets.

\textbf{CIFAR-10}
\underline{(i) Model.} We use ResNet-20 \citep{he2016deep}, RepVGG-A0 \citep{ding2021repvgg}, and VGG-11 \citep{simonyan2014very}. They are trained on CIFAR-10 training set.
\underline{(ii) Synthetic Shift.} Similar to ImageNet-C, we use CIFAR-10-C \citep{hendrycks2019robustness} to study the synthetic shift.
It contains $19$ types of corruption and each type has $5$ intensity levels.
\underline{(iii) Real-world Shift.} We include three test sets: 1) CIFAR-10.1 with reproduction shift  \citep{recht2018cifar}, 2) CIFAR-10.2 with reproduction shift~\citep{recht2018cifar}, and 3) CINIC-10 that is sampled from a different database ImageNet.

\textbf{CUB-200.} We also consider fine-grained categorization with large intra-class variations and small inter-class variations~\citep{wei2021fine}.
We build up a setup based on CUB-200-2011~\citep{wah2011caltech} that contains 200 birds categories.
\underline{(i) Model.} We use $3$ classifiers: ResNet-50,  ResNet-101, and PMG \citep{du2020fine}. They are pretrained on ImageNet and finetuned on CUB-200-2011 training set. We use the publicly available codes provided by \citep{du2020fine}.
\underline{(ii) Synthetic Shift.} Following the protocol in ImageNet-C, we create CUB-200-C by applying $19$ types of corruptions with $5$ intensity levels to CUB-200-2011 test set.
\underline{(iii) Real-world Shift.} We use CUB-200-P(aintings) with style shift \citep{wang2020progressive}. It contains bird paintings with various rendition (\textit{e.g.,} watercolors, oil paintings, pencil drawings, stamps, and cartoons) collected from web.

\subsection{Compared Methods and Evaluation Metrics}
We use \textbf{four} existing measures for comparison. They are all developed based on the softmax output of classifier. %
\textbf{1)} \emph{Average Confidence (AC)} \citep{hendrycks2017baseline}. The average of maximum softmax scores on the target dataset;
\textbf{2)} \emph{Average Negative Entropy (ANE)} \citep{guillory2021predicting}. The average of negative entropy scores on the target dataset;
\textbf{3)} \emph{Average Thresholded Confidence (ATC)} \citep{garg2022leveraging}. This method first identifies a threshold on source validation set. Then, ATC is defined as the expected number of target images that obtain a softmax confidence score than the threshold;
\textbf{4)} \emph{Difference of Confidence (DOC)} \citep{guillory2021predicting}. It is defined as the source validation accuracy minus the difference of AC on the target dataset and source validation set. The difference of AC is regarded as a surrogate of distribution shift.

\begin{table*}[t]
	\begin{center}
	 \caption{\textbf{Method comparison under ImageNet, CIFAR-10, and CUB-200 setups}. We compare nuclear norm with four existing methods. To quantify the effectiveness in assessing OOD generalization, we report coefficients of determination ($R^2$) and Spearman's rank correlation ($\rho$). The highest score in each row is highlighted in \textbf{bold}. We show that nuclear norm exhibits the highest correlation strength ($R^2$ and $\rho$) with OOD accuracy across three setups. 
	}\label{tab:corr}
	\vspace{2pt}
	\setlength{\tabcolsep}{8pt}
	\small
	\begin{tabular}{lc cc cc cc cc cc}
	\toprule
\multirow{2}{*}{\textbf{Setup}}  & \multirow{2}{*}{\textbf{Model}}  & \multicolumn{2}{c}{\textbf{AC}} & \multicolumn{2}{c}{\textbf{ANE}} & \multicolumn{2}{c}{\textbf{ATC}} & \multicolumn{2}{c}{\textbf{DoC}} & \multicolumn{2}{c}{\textbf{Nuclear Norm}}\\
\cmidrule(lr){3-4}  \cmidrule(lr){5-6} \cmidrule(lr){7-8} \cmidrule(lr){9-10}  \cmidrule(lr){11-12} 
       &  & $R^2$ & $\rho$ & $R^2$ & $\rho$ & $R^2$ & $\rho$ & $R^2$ & $\rho$ & $R^2$ & $\rho$ \\
		\midrule
		\multicolumn{1}{l}{\multirow{7}{*}{\makecell{ImageNet}}} 
		& ViT & 0.970 & 0.990 %
		      & 0.964 & 0.988 %
			  & 0.978 & 0.990 %
		      & 0.961 & 0.990 %
		      &  \textbf{0.991}  &  \textbf{0.995}  \\ 
		& BeiT 
		      &  0.977 & 0.994  %
		      &  0.964 & 0.989 %
	          &  0.985 & {0.995} %
              &  0.979 & 0.994 %
		      &  \textbf{0.988}  &  \textbf{0.996}  \\ 
		& Swin 
		      &  0.794   &  0.929  %
		      &  0.732  &  0.909 %
			  &  0.815  &  0.935 %
		      &  0.791  &  0.929 %
		      &  \textbf{0.949}  &  \textbf{0.961}  \\ 
		& DenseNet 
		      &  0.938  &  0.984 %
		      &  0.929  &  0.979 %
			  &  0.961  &  0.989 %
		      &  0.937  &  0.984 %
		      &  \textbf{0.995}  &  \textbf{0.997}  \\ 
		& Res152-BiT 
		      &  0.891  &  0.981 %
		      &  0.877  &  0.979 %
			  &  0.916  &  0.982 %
		      &  0.908  &  0.981 %
		      &  \textbf{0.981}  &  \textbf{0.991}  \\ 
		& ConvNeXt 
		      &  0.894  &  0.971 %
		      &  0.866  &  0.960 %
		      &  0.888  &  0.967 %
		      &  0.899  &  0.971 %
		      &  \textbf{0.967}  &  \textbf{0.982}  \\ 
		\cmidrule(lr){2-12}
		& \cellcolor{mygray}{Average}    
		& \cellcolor{mygray}{0.911}  & \cellcolor{mygray}{0.975} %
         & \cellcolor{mygray}{0.889} & \cellcolor{mygray}{0.968} %
         & \cellcolor{mygray}{0.924} & \cellcolor{mygray}{0.976} %
         & \cellcolor{mygray}{0.911} & \cellcolor{mygray}{0.975} %
         & \cellcolor{mygray}{\textbf{0.979}} & \cellcolor{mygray}{\textbf{0.989}} \\ 
	\midrule
		\multicolumn{1}{l}{\multirow{4}{*}{{CIFAR-10}}}
		& ResNet-20 & 0.916 & 0.991 %
		            & 0.916 & 0.991 %
		            & 0.934 & 0.992 %
		            & 0.937 & 0.991 %
		            & \textbf{0.989} & \textbf{0.995}\\ 
		& RepVGG-A0 & 0.811 & 0.982 %
		            & 0.806 & 0.981 %
		            & 0.841 & 0.985 %
		            & 0.824 & 0.982 %
		            & \textbf{0.992} & \textbf{0.996}\\ 
		& VGG-11   & 0.973  & {0.994} %
		           & 0.973  & 0.995 %
		           & 0.984  & \textbf{0.996} %
		           & 0.964  & {0.994} %
		           & \textbf{0.988} & \textbf{0.996}\\ 
		\cmidrule(lr){2-12}
		& \cellcolor{mygray}{Average}    
         & \cellcolor{mygray}{0.900} & \cellcolor{mygray}{0.989}%
         & \cellcolor{mygray}{0.900} & \cellcolor{mygray}{0.988} %
         & \cellcolor{mygray}{0.920} & \cellcolor{mygray}{0.991} %
         & \cellcolor{mygray}{0.908} & \cellcolor{mygray}{0.989} %
         & \cellcolor{mygray}{\textbf{0.990}} & \cellcolor{mygray}{\textbf{0.995}} \\  
	\midrule
		\multicolumn{1}{l}{\multirow{4}{*}{{CUB-200}}}
		& ResNet-50  & 0.836 & 0.942 %
		             & 0.839 & 0.939 %
		             & 0.855 & 0.957 %
		             & 0.818 & 0.942 %
		             & \textbf{0.989} & \textbf{0.997} \\ 
		& ResNet-101 & 0.303 & 0.734 %
		             & 0.319 & 0.739 %
		             & 0.351 & 0.775 %
		             & 0.308 & 0.734 %
		             & \textbf{0.987} & \textbf{0.998} \\
		& PMG        & 0.892 & 0.979 %
		             & 0.893 & 0.977 %
		             & 0.977 & 0.991 %
		             & 0.903 & 0.979 %
		             & \textbf{0.990} & \textbf{0.998} \\ 
		\cmidrule(lr){2-12}
		& \cellcolor{mygray}{Average}
         & \cellcolor{mygray}{0.677} & \cellcolor{mygray}{0.885} %
         & \cellcolor{mygray}{0.684} & \cellcolor{mygray}{0.885}  %
         & \cellcolor{mygray}{0.727} & \cellcolor{mygray}{0.908}  %
         & \cellcolor{mygray}{0.677} & \cellcolor{mygray}{0.885} %
         & \cellcolor{mygray}{\textbf{0.989}} & \cellcolor{mygray}{\textbf{0.997}} \\ 
		\bottomrule
	\end{tabular}
	\end{center}
	\vspace{-10pt}
\end{table*}

\textbf{Evaluation Procedure.} Given a trained classifier, we test it on $95$ synthesized test sets under each setup. For each test set, we calculate the ground-truth accuracy and the estimated OOD quantity. Then, we evaluate the correlation strength between the estimated OOD quantity and accuracy. We also show scatter plots and mark real-world datasets to compare different approaches. 

\textbf{Evaluation Metrics.} To measure the quality of estimations, we use Pearson Correlation coefficient ($r$) \citep{benesty2009pearson} and Spearman's Rank Correlation coefficient ($\rho$) \citep{kendall1948rank} to quantify the linearity and monotonicity.
They range from $[-1, 1]$. A value closer to $1$ (or $-1$) indicates strong positive (or negative) correlation, and $0$ implies no correlation \cite{benesty2009pearson}.
To precisely show the correlation, we use prob axis scaling that maps the range of both accuracy and estimated OOD quantity from $[0,1]$ to $[-\infty, +\infty]$, following \cite{taori2020measuring, miller2021accuracy}. 
We also report the coefficient of determination ($R^2$) \citep{nagelkerke1991note} of the linear fit between estimated OOD quantity and accuracy following \citep{yu2022predicting}. The coefficient $R^2$ ranges from $0$ to $1$. An $R^2$ of $1$ indicates that regression predictions perfectly fit OOD accuracy.

\subsection{Main Results}

\textbf{Nuclear norm is an effective indicator to OOD accuracy.} In Table \ref{tab:corr}, we report the correlation results of nuclear norm under three setups: ImageNet-1k, CIFAR-10, and CUB-200. 
We consistently observe a very strong correlation ($R^2>0.945$ and $\rho>0.960$) between the nuclear norm and ODD accuracy under the three setups. The strong correlation still exists when using different model architectures under each setup. 
For example, the average coefficients of determination $R^2$ achieved by nuclear norm are $0.979$, $0.990$, and $0.989$ on ImageNet-1k, CIAFR-10, and CUB-200, respectively. It demonstrates that nuclear norm well captures the distribution shift and makes excellent OOD accuracy estimations for different classifiers.

\textbf{Nuclear norm is generally more robust and accurate than existing methods.} 
Compared with existing methods, nuclear norm achieves the strongest correlation with classifier performance across all three setups.
With different models on ImageNet, nuclear norm achieves an average $R^2$ of $0.979$, while the second best method (ATC) only obtains~$0.924$. Moreover, nuclear norm outperforms ATC by $0.262$ and in average $R^2$ under CUB-200 setup.
We note that the prediction performance of nuclear norm is overall more robust than other methods. Competing methods are less effective in predicting the accuracy of certain classifiers such as Swin under the ImageNet setup and ResNet-101 under the CUB-200 setup. 
For these difficult cases, nuclear norm remains useful and effective with $R^2>0.945$.

\begin{figure*}[t]
    \begin{center}
    \includegraphics[width=0.9\linewidth]{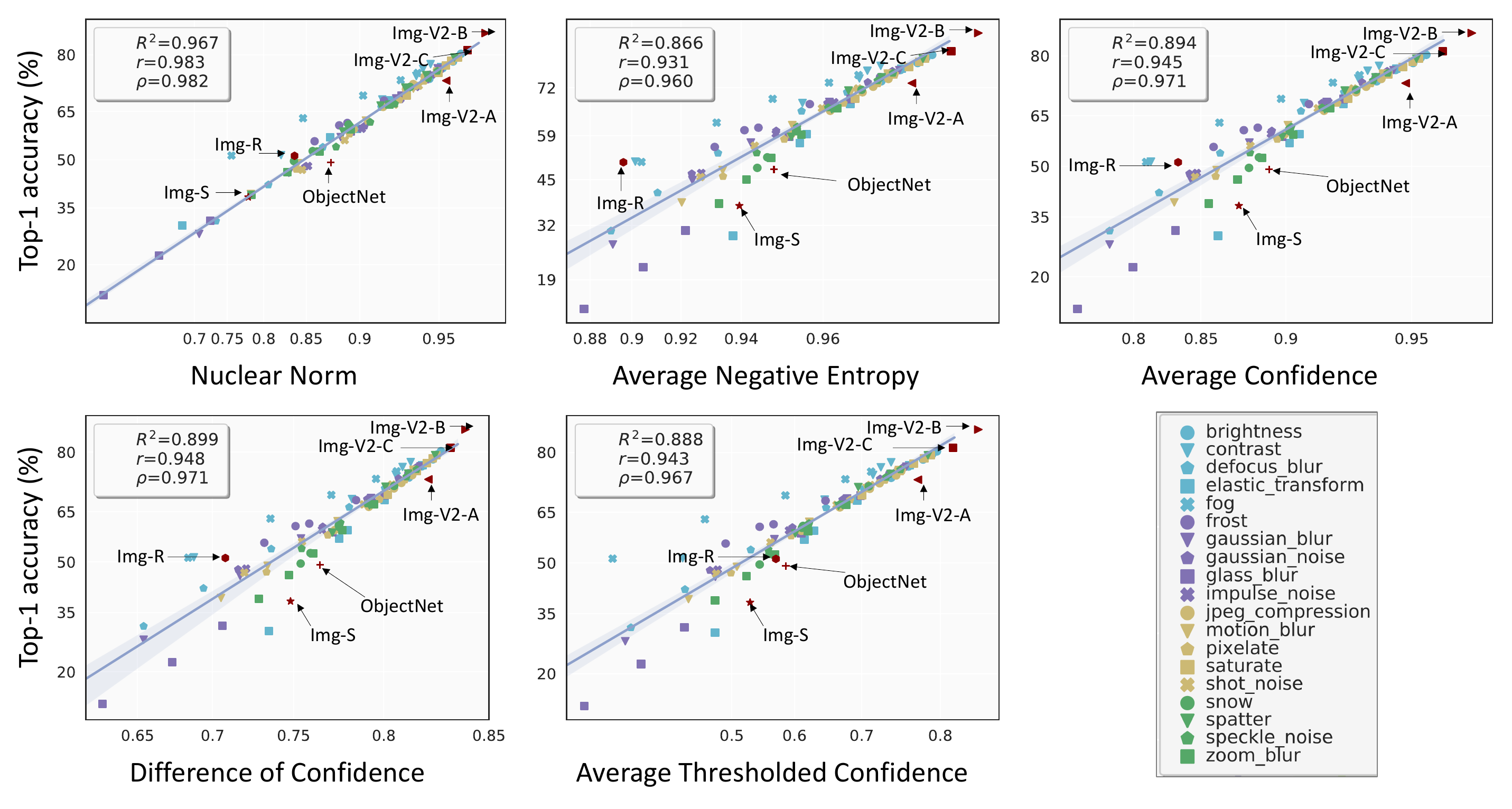}
        \vspace{-12pt}
    \caption{\textbf{Correlation study under the ImageNet setup.} We plot the actual accuracy of \textbf{\textit{ConvNeXt}} and \textcolor{black}{five measures including nuclear norm and four competing methods}. 
    Different shapes in each sub-figure represent different test sets. The straight lines are calculated by linear regression fit on synthetic datasets of ImageNet-C. 
    We list the $19$ types of corruptions in ImageNet-C using different shapes and colors in the bottom right figure.
    We also mark the $6$ real-world datasets in each sub-figure with arrows.
    We observe nuclear norm exhibits stronger correlation with accuracy. Moreover, with nuclear norm, real-world test sets are closely around the linearly fit line.
    }\label{fig:img}
    \end{center}
    \vspace{-11pt}
\end{figure*}

\begin{figure*}[!ht]
    \begin{center}
    \includegraphics[width=1\linewidth]{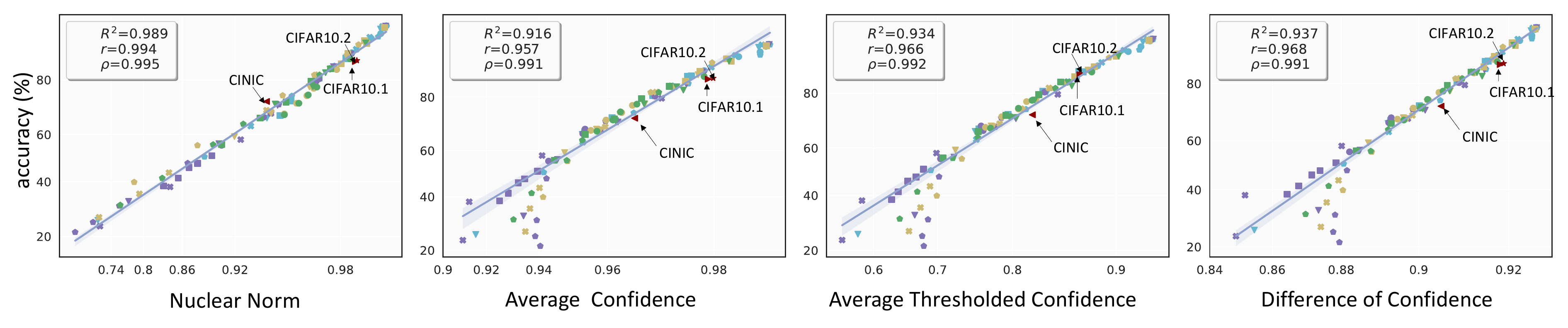}
     \vspace{-18pt}
    \caption{\textbf{Correlation study under the CIFAR-10 setup.} We plot the actual accuracy of \textbf{\textit{ResNet-20}} and the estimated OOD quantity. We show the results of nuclear norm, AC and ATC.
   The lines are calculated by linear regression fit on CIFAR-C. 
    We mark the $3$ real-world test sets in each sub-figure.
    We show that AC and ATC fail to estimate generalization on datasets with lower ground-truth accuracy. In comparison, nuclear norm is more robust and accurate.
    }\label{fig:cifar}
    \end{center}
     \vspace{-10pt}
\end{figure*}

\begin{figure*}[t]
    \begin{center}
    \includegraphics[width=1\linewidth]{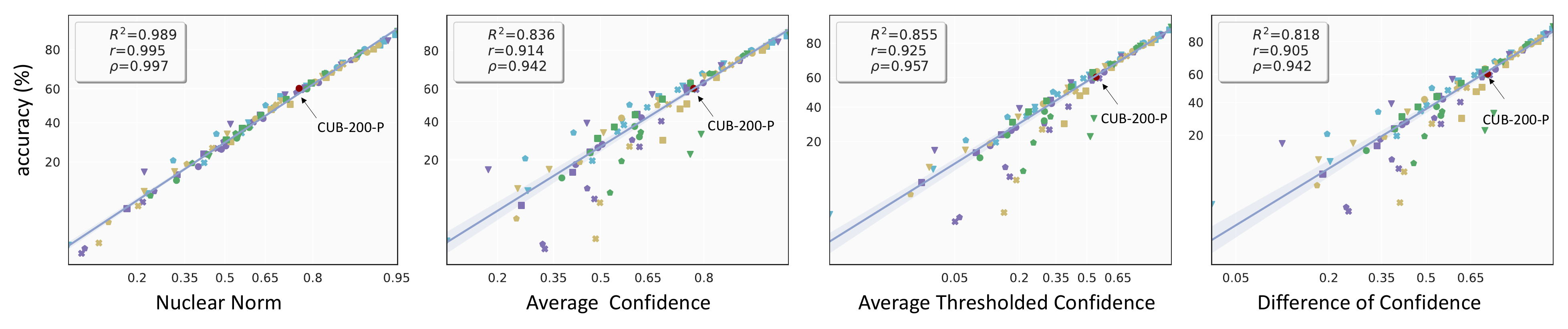}
    \vspace{-0.9cm}
    \caption{\textbf{Correlation study under the CUB-200 setup.}
    We plot the actual accuracy of \textbf{\textit{ResNet-50}} and the estimated OOD quantity. We compare nuclear norm with AC and ATC.
    The straight lines are calculated by the linear regression fit on CUB-200-C. 
    We mark the real-world test set CUB-P in each sub-figure.
    While AT and ATC cannot give accurate estimates for some datasets, nuclear norm is more robust and accurate in predicting generalization. Specifically, all test sets are closely around the line, yielding higher correlation strength.
    \vspace{-0.4cm}
    }\label{fig:cub}
    \end{center}
\end{figure*}

\textbf{Nuclear norm can estimate the accuracy of real-world datasets}. To further validate the effectiveness of nuclear norm, we show its accuracy prediction on real-world datasets as the scatter plots under the three setups (Fig.~\ref{fig:img}, Fig.~\ref{fig:cifar}, and Fig.~\ref{fig:cub}, respectively). 
We observe that nuclear norm can produce reasonably accurate estimates on real-world test sets.
Under the ImageNet setup (Fig.~\ref{fig:img}), the six test sets (\textit{e.g.}, ImageNet-V2/A/B/C and ImageNet-R) are very close to the linear regression line. It demonstrates that nuclear norm well captures these real-world shifts and thus estimates OOD performance very well. 
Under CIFAR-10 and CUB-200 setups, we have similar observations.

Although existing methods (\textit{e.g.}, ATC) are effective on most real-world datasets, nuclear norm still shows its advantage over them. Other methods fail to capture the shifts of ImageNet-S and ObjectNet under the ImageNet setup: they are far away from lines.
In comparison, nuclear norm captures them well and both datasets are very close to lines. Furthermore, the scatter plots under CIFAR-10 (Fig.~\ref{fig:cifar}) and CUB-200 (Fig.~\ref{fig:cub}) show that the competing methods often give accuracy numbers lower than the ground truth when the test set is difficult, while the nuclear norm is still effective.

\subsection{Discussion and Analysis}\label{sec:discussion}

\begin{figure*}[t]
    \begin{center}
    \includegraphics[width=1\linewidth]{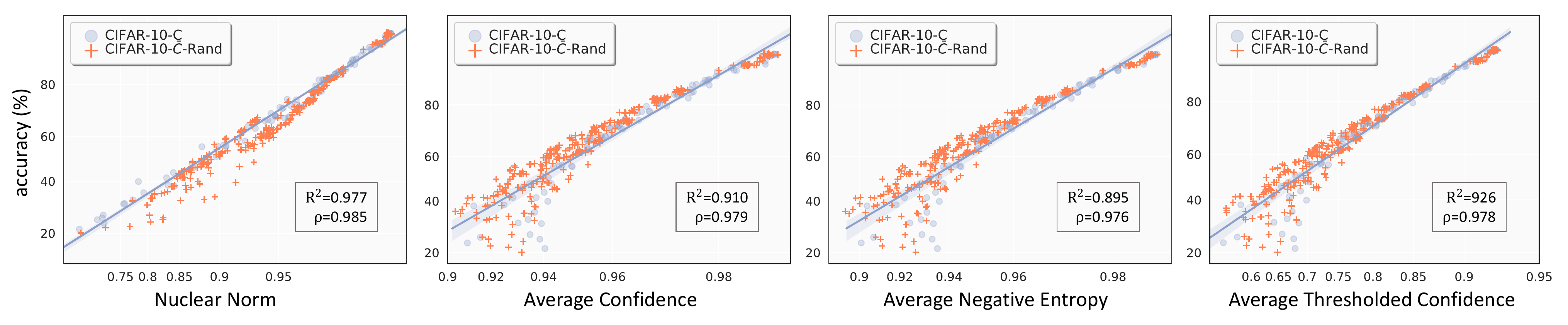}
    \vspace{-0.8cm}
    \caption{\textbf{\wj{Correlation study on randomly synthesized datasets under the CIFAR-10 setup.}} We report results with ResNet-20.
    Randomly synthesized datasets (CIFAR-10-$\bar{C}$-Rand) are marked with orange ``$+$", and the solid lines are fit with robust linear regression on controllable CIFAR-10-C. 
    Overall, CIFAR-10-$\bar{C}$-Rand datasets are distributed around the line for every method.
    Looking more closely at the low-accuracy region (bottom left in each subfigure), nuclear norm is more effective than other methods.
    \vspace{-0.4cm}
    }\label{fig:rand}
    \end{center}
\end{figure*}

\textbf{(I) Beyond controllable synthetic shifts.}
The synthetic datasets (\textit{e.g.}, ImageNet-C) are algorithmically generated in a controllable manner. 
Here, we investigate whether a measure is robust in predicting OOD accuracy on random synthetic datasets. To this end, we randomly synthesize datasets for the \textit{CIFAR-10 setup}. Specifically, we use $10$ new corruptions of ImageNet-$\bar{C}$~\citep{mintun2021interaction} that are \textit{perceptually dissimilar} to ImageNet-C. The dissimilar corruptions include warps, blurs, color distortions, noise additions, and obscuring effects.
When synthesizing each test set, we randomly choose $3$ corruptions and make corruption strength random. By doing so, we create $200$ random synthetic datasets denoted CIFAR-$\bar{C}$-Rand. 

In Fig. \ref{fig:rand}, we report the correlation results using ResNet-20 under the CIFAR-10 setup. We also show the linear regression lines that fit on datasets of CIFAR-10-C. 
We report the results of four methods including nuclear norm, AC, ATC, and DoC. 
We have two observations. First, for each method, CIFAR-$\bar{C}$-Rand datasets (marked with ``$+$") are generally distributed around the linear lines. This indicates that all methods can make reasonable accuracy estimations on CIFAR-$\bar{C}$-Rand. 
Second, for the low-accuracy region (bottom left in each subfigure), nuclear norm gives more accurate and robust predictions than other methods. 

\textbf{(II) \wj{Other measures to consider prediction confidence and dispersity.}} 
\wj{Here, we discuss the usage of other measures. We study 
mutual information maximizing (MI) which is commonly used in discriminative clustering} \citep{bridle1991unsupervised,krause2010discriminative}.
Recent methods use it as a regularization to make model predictions confident and diverse \citep{liang2020we,yang2021exploiting,tang2020unsupervised}.
Given a prediction matrix  $\mP\in\mathbb{R}^{n_t \times k}$, IM is defined as $H(\frac{1}{n_t}\sum_{i=1}^{n_t}{\mP_{i,:}}) - \frac{1}{n_t}\sum_{i=1}^{n_t}H(\mP_{i,:})$.
Its first term encourages the predictions to be globally balanced. The second term is standard entropy that makes the prediction confident.
In Table \ref{tab:im}, we report the correlation results using MI. 
We observe that MI and nuclear norm achieve similar average correlation strength.
Compared with average negative entropy (ANE), MI exhibits stronger correlation across three setups. For example, MI yields a $0.110$ higher $\rho$ than ANE on CUB. 
{This further validates that prediction dispersity is informative for accuracy estimation.}

\begin{figure*}[!ht]
    \begin{center}
    \includegraphics[width=1\linewidth]{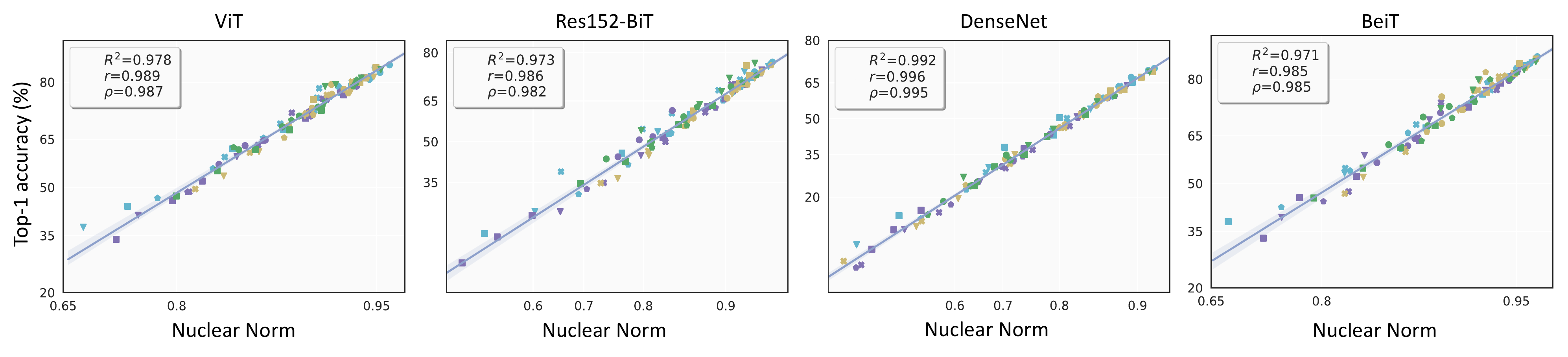}
    \vspace{-0.7cm}
    \caption{\textbf{Analysis of the influence of test set size on nuclear norm.} We conduct correlation study on \textit{randomly sub-sampled} ImageNet-C. Specifically, we vary the size of each dataset by randomly selecting $20$--$90\%$ of test samples.
    We test three classifiers and observe the correlation strength remains very high ($R^2>0.960$ and $\rho>0.970$).
    }\label{fig:size}
     \vspace{-8pt}
    \end{center}
\end{figure*}

 \textbf{III. Impact of test set size.}
 As illustrated in Section \ref{sec:nuclear_norm}, nuclear norm \textcolor{black}{without scaling} is related to the size of the prediction matrix. Since test sets can contain any number of data points, we normalize nuclear norm by its upper bound to make it robust to test set size.
Here, we change the size of each dataset of ImageNet-C by randomly selecting $20$--$90\%$ of all test samples.
As shown in Fig. \ref{fig:size}, scaled nuclear norm is well correlated with accuracy with different dataset sizes. 

\begin{table}[t]
	\setlength{\tabcolsep}{4pt}
    \begin{tabular}{c c c c}
        \toprule
         \textbf{Method} &  \multicolumn{1}{c}{\textbf{ImageNet-1k}} & \multicolumn{1}{c}{\textbf{CIFAR-10}} & \multicolumn{1}{c}{\textbf{CUB-200}}\\ 
        \midrule
         ANE & 0.968 & 0.988 & 0.885  \\
         \midrule
         MI & 0.982 & 0.994 & 0.995  \\
         Nuclear Norm & 0.989  & 0.995 & 0.997\\
        \bottomrule
    \end{tabular}
    \caption{\textbf{Correlation results using mutual information maximizing (MI).} We report the average correlation strength (Spearman's rank correlation $\rho$) under each setup.
    We observe MI and nuclear norm have similar correlation strengths. Compared with average negative entropy (ANE), MI exhibits a stronger correlation with accuracy across three setups. This also indicates that the prediction dispersity is informative for accuracy estimation.
    } \label{tab:im}
    \vspace{-0.5cm}
\end{table}

\begin{figure*}[t]
    \begin{center}
    \includegraphics[width=0.9\linewidth]{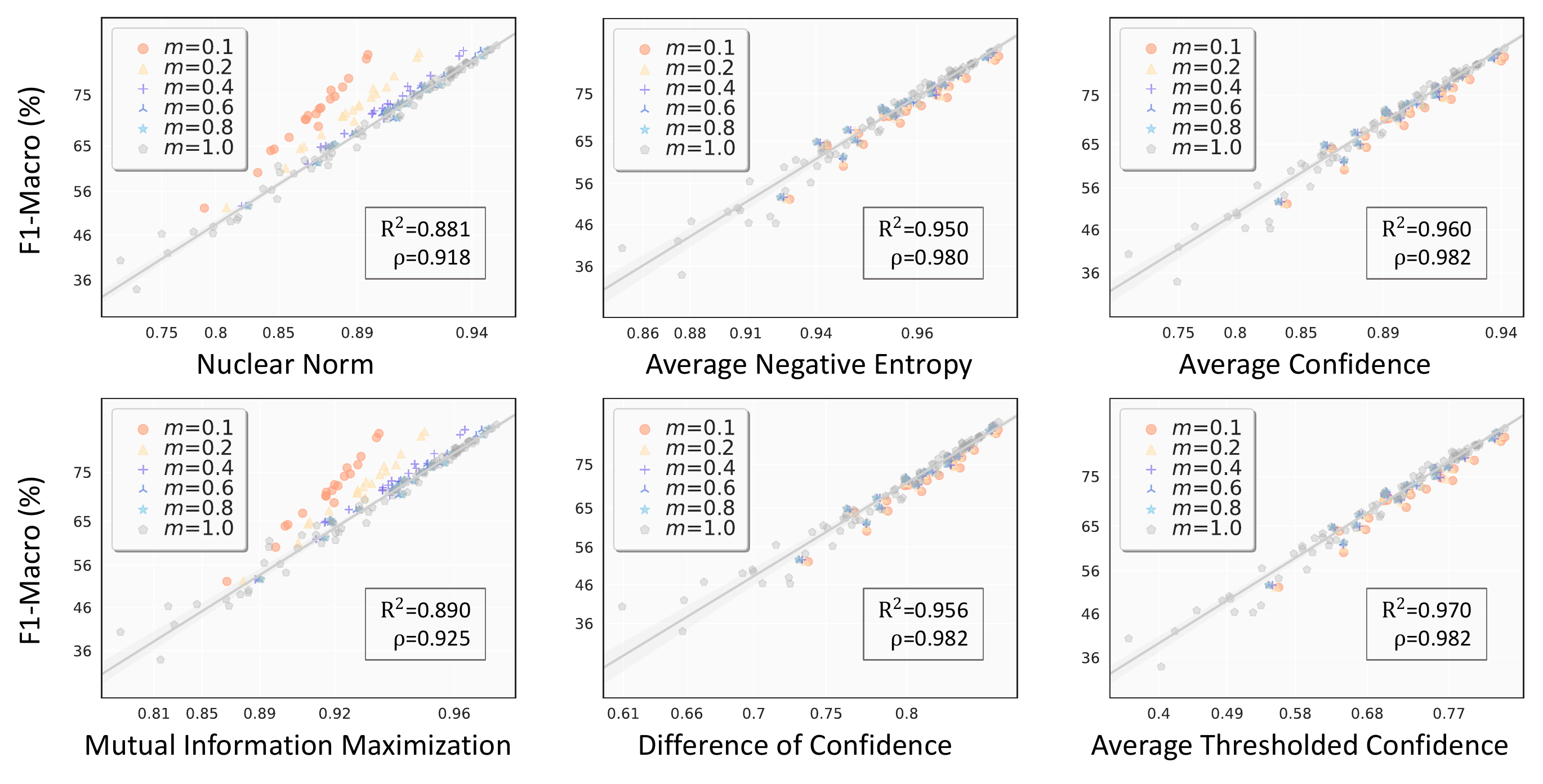}
    \vspace{-0.4cm}
    \caption{\wj{\textbf{Comparison of various methods on imbalanced test sets.} 
    Using \textbf{ViT} under ImageNet setup, we study the robustness of existing methods to several imbalance ratio $m$ when test sets are long-tailed. 
    A smaller $m$ indicates a higher imbalance intensity. The linear lines are fit on standard test sets ($m=1$). 
    We observe that both mutual information maximization (MI) and nuclear norm are less effective than other methods under strong-imbalanced datasets ($m<0.4$). Furthermore, we show that MI and nuclear norm are robust under mild-imbalanced test sets ($m\geq0.4$).}
    \vspace{-0.35cm}
    }\label{fig:imb}
    \end{center}
\end{figure*}

\textbf{(IV) Discussion on label shift (class imbalance).}
In our work, we consider the common covariate shift \citep{sugiyama2012machine} where $p_{S}(\vx)\neq p_{T}(\vx)$ and $p_{S}(y|\vx)=p_{T}(y|\vx)$ (\emph{i.e.}, the class label of the input data is independent of distribution).
Nuclear norm measures the prediction dispersity and thus implicitly assumes that the test set \textit{does not contain strong} label shift (\emph{i.e.}, class imbalance). As for the label shift~\citep{garg2020unified}, the assumption about the distribution is $p_{S}(y) \neq p_{T}(y)$ and $p_{S}(\vx|y)=p_{T}(\vx|y)$ (\emph{i.e.}, the class-conditional distribution does not change).

Here, we discuss the robustness of nuclear norm to label shift. We first note that real-world test sets such as ImageNet-R, ObjectNet, and CUB-200-P are already imbalanced. 
We show that nuclear norm robustly captures them: they are very close to the linear lines (as shown in Fig.~\ref{fig:img} and Fig.~\ref{fig:cub})
\wj{To further study the effect of label shift, we create long-tailed imbalance test sets. We use exponential decay~\cite{cao2019learning} to make the proportion of each class different.}
We use an imbalance ratio $m$ to denote the ratio between sample sizes of the least frequent and most frequent classes. We test several imbalanced ratios: $\{0.1, 0.2, 0.4, 0.6, 0.8\}$. 
We conduct experiments on ImageNet-C and use $19$ types of corruption datasets with the second intensity level.
\wj{As shown in Fig. \ref{fig:imb}, we observe that both nuclear norm and MI are influenced by label shift when the imbalance is strong ($m<=0.2$).}
For example, when the test set is of extreme class imbalance ($m=0.1$), the prediction of nuclear norm is not accurate.
We also observe that under the strong imbalance ($m<=0.2$), exiting methods (\emph{e.g.}, ATC) are more stable than nuclear norm and MI.
We note that both nuclear norm and MI are robust in the presence of moderate label shift($m\geq0.4$).

We further emphasize that considering the prediction dispersity under severe class imbalance remains useful. Specifically, if we have prior knowledge about the long-tailed class distribution, we can expect class predictions to follow it rather than a uniform distribution. In this way, we can more accurately characterize the class-specific prediction dispersity for the task of accuracy estimation. For example, modifying the second term of MI would be helpful. That said, it is a potential research direction to further study this idea by considering extra techniques such as label shift estimation~\citep{lipton2018detecting,tian2020posterior} and prior knowledge~\citep{chen2021mandoline,sun2022prior}.

\section{Conclusion}
This work studies accuracy estimation  where the goal is to predict classifier accuracy on unlabeled test sets. 
While existing methods study the confidence of prediction matrix on unlabelled data, this work proposes to consider prediction dispersity.
We first show that prediction dispersity is a useful property that correlates strongly with classifier accuracy on various test sets.
Then, we consider both prediction confidence and dispersity using nuclear norm to achieve more accurate predictions.
Across three setups, we consistently observe that nuclear norm is more effective and robust in assessing classifier OOD performance than existing methods. We further conduct experiments on imbalanced test sets and show that nuclear norm is still effective under moderate class imbalances. Finally, we study its limitation under severe class imbalance and discuss potential solutions.
\newpage

\bibliography{icml}

\begin{thebibliography}{76}
\providecommand{\natexlab}[1]{#1}
\providecommand{\url}[1]{\texttt{#1}}
\expandafter\ifx\csname urlstyle\endcsname\relax
  \providecommand{\doi}[1]{doi: #1}\else
  \providecommand{\doi}{doi: \begingroup \urlstyle{rm}\Url}\fi

\bibitem[Arora et~al.(2018)Arora, Ge, Neyshabur, and Zhang]{arora2018stronger}
Arora, S., Ge, R., Neyshabur, B., and Zhang, Y.
\newblock Stronger generalization bounds for deep nets via a compression
  approach.
\newblock \emph{arXiv preprint arXiv:1802.05296}, 2018.

\bibitem[Baek et~al.(2022)Baek, Jiang, Raghunathan, and
  Kolter]{baek2022agreement}
Baek, C., Jiang, Y., Raghunathan, A., and Kolter, Z.
\newblock Agreement-on-the-line: Predicting the performance of neural networks
  under distribution shift.
\newblock \emph{arXiv preprint arXiv:2206.13089}, 2022.

\bibitem[Baldock et~al.(2021)Baldock, Maennel, and Neyshabur]{baldock2021deep}
Baldock, R., Maennel, H., and Neyshabur, B.
\newblock Deep learning through the lens of example difficulty.
\newblock \emph{Advances in Neural Information Processing Systems},
  34:\penalty0 10876--10889, 2021.

\bibitem[Barbu et~al.(2019)Barbu, Mayo, Alverio, Luo, Wang, Gutfreund,
  Tenenbaum, and Katz]{barbu2019objectnet}
Barbu, A., Mayo, D., Alverio, J., Luo, W., Wang, C., Gutfreund, D., Tenenbaum,
  J., and Katz, B.
\newblock Objectnet: A large-scale bias-controlled dataset for pushing the
  limits of object recognition models.
\newblock In \emph{Advances in neural information processing systems}, 2019.

\bibitem[Benesty et~al.(2009)Benesty, Chen, Huang, and
  Cohen]{benesty2009pearson}
Benesty, J., Chen, J., Huang, Y., and Cohen, I.
\newblock Pearson correlation coefficient.
\newblock In \emph{Noise reduction in speech processing}, pp.\  1--4. Springer,
  2009.

\bibitem[Bridle et~al.(1991)Bridle, Heading, and
  MacKay]{bridle1991unsupervised}
Bridle, J., Heading, A., and MacKay, D.
\newblock Unsupervised classifiers, mutual information and'phantom targets.
\newblock \emph{Advances in neural information processing systems}, 4, 1991.

\bibitem[Cao et~al.(2019)Cao, Wei, Gaidon, Arechiga, and Ma]{cao2019learning}
Cao, K., Wei, C., Gaidon, A., Arechiga, N., and Ma, T.
\newblock Learning imbalanced datasets with label-distribution-aware margin
  loss.
\newblock \emph{Advances in neural information processing systems}, 32, 2019.

\bibitem[Chen et~al.(2021{\natexlab{a}})Chen, Liu, Avci, Wu, Liang, and
  Jha]{chen2021detecting}
Chen, J., Liu, F., Avci, B., Wu, X., Liang, Y., and Jha, S.
\newblock Detecting errors and estimating accuracy on unlabeled data with
  self-training ensembles.
\newblock \emph{Advances in Neural Information Processing Systems}, 34,
  2021{\natexlab{a}}.

\bibitem[Chen et~al.(2021{\natexlab{b}})Chen, Goel, Sohoni, Poms, Fatahalian,
  and R{\'e}]{chen2021mandoline}
Chen, M., Goel, K., Sohoni, N.~S., Poms, F., Fatahalian, K., and R{\'e}, C.
\newblock Mandoline: Model evaluation under distribution shift.
\newblock In \emph{International Conference on Machine Learning}, pp.\
  1617--1629, 2021{\natexlab{b}}.

\bibitem[Chrabaszcz et~al.(2017)Chrabaszcz, Loshchilov, and
  Hutter]{chrabaszcz2017downsampled}
Chrabaszcz, P., Loshchilov, I., and Hutter, F.
\newblock A downsampled variant of imagenet as an alternative to the cifar
  datasets.
\newblock \emph{arXiv preprint arXiv:1707.08819}, 2017.

\bibitem[Chuang et~al.(2020)Chuang, Torralba, and
  Jegelka]{chuang2020estimating}
Chuang, C.-Y., Torralba, A., and Jegelka, S.
\newblock Estimating generalization under distribution shifts via
  domain-invariant representations.
\newblock In \emph{International Conference on Machine Learning}, 2020.

\bibitem[Chuang et~al.(2021)Chuang, Mroueh, Greenewald, Torralba, and
  Jegelka]{chuang2021measuring}
Chuang, C.-Y., Mroueh, Y., Greenewald, K., Torralba, A., and Jegelka, S.
\newblock Measuring generalization with optimal transport.
\newblock In \emph{Advances in Neural Information Processing Systems},
  volume~34, pp.\  8294--8306, 2021.

\bibitem[Corneanu et~al.(2020)Corneanu, Escalera, and
  Martinez]{corneanu2020computing}
Corneanu, C.~A., Escalera, S., and Martinez, A.~M.
\newblock Computing the testing error without a testing set.
\newblock In \emph{Proceedings of the IEEE conference on computer vision and
  pattern recognition}, pp.\  2677--2685, 2020.

\bibitem[Cui et~al.(2020)Cui, Wang, Zhuo, Li, Huang, and Tian]{cui2020towards}
Cui, S., Wang, S., Zhuo, J., Li, L., Huang, Q., and Tian, Q.
\newblock Towards discriminability and diversity: Batch nuclear-norm
  maximization under label insufficient situations.
\newblock In \emph{Proceedings of the IEEE/CVF Conference on Computer Vision
  and Pattern Recognition}, pp.\  3941--3950, 2020.

\bibitem[Cui et~al.(2021)Cui, Wang, Zhuo, Li, Huang, and Tian]{cui2021fast}
Cui, S., Wang, S., Zhuo, J., Li, L., Huang, Q., and Tian, Q.
\newblock Fast batch nuclear-norm maximization and minimization for robust
  domain adaptation.
\newblock \emph{arXiv preprint arXiv:2107.06154}, 2021.

\bibitem[Deng et~al.(2009)Deng, Dong, Socher, Li, Li, and
  Fei-Fei]{deng2009imagenet}
Deng, J., Dong, W., Socher, R., Li, L.-J., Li, K., and Fei-Fei, L.
\newblock Imagenet: A large-scale hierarchical image database.
\newblock In \emph{Proceedings of the IEEE Conference on Computer Vision and
  Pattern Recognition}, pp.\  248--255, 2009.

\bibitem[Deng \& Zheng(2021)Deng and Zheng]{deng2021labels}
Deng, W. and Zheng, L.
\newblock Are labels always necessary for classifier accuracy evaluation?
\newblock In \emph{Proceedings of the IEEE/CVF Conference on Computer Vision
  and Pattern Recognition}, pp.\  15069--15078, 2021.

\bibitem[Deng et~al.(2021)Deng, Gould, and Zheng]{Deng:ICML2021}
Deng, W., Gould, S., and Zheng, L.
\newblock What does rotation prediction tell us about classifier accuracy under
  varying testing environments?
\newblock In \emph{International conference on machine learning}, 2021.

\bibitem[Deng et~al.(2022)Deng, Gould, and Zheng]{deng2022strong}
Deng, W., Gould, S., and Zheng, L.
\newblock On the strong correlation between model invariance and
  generalization.
\newblock In \emph{Advances in Neural Information Processing Systems}, 2022.

\bibitem[Ding et~al.(2021)Ding, Zhang, Ma, Han, Ding, and Sun]{ding2021repvgg}
Ding, X., Zhang, X., Ma, N., Han, J., Ding, G., and Sun, J.
\newblock Repvgg: Making vgg-style convnets great again.
\newblock In \emph{Proceedings of the IEEE/CVF Conference on Computer Vision
  and Pattern Recognition}, pp.\  13733--13742, 2021.

\bibitem[Donmez et~al.(2010)Donmez, Lebanon, and
  Balasubramanian]{donmez2010unsupervised}
Donmez, P., Lebanon, G., and Balasubramanian, K.
\newblock Unsupervised supervised learning i: Estimating classification and
  regression errors without labels.
\newblock \emph{Journal of Machine Learning Research}, 11\penalty0 (4), 2010.

\bibitem[Dosovitskiy et~al.(2020)Dosovitskiy, Beyer, Kolesnikov, Weissenborn,
  Zhai, Unterthiner, Dehghani, Minderer, Heigold, Gelly,
  et~al.]{dosovitskiy2020image}
Dosovitskiy, A., Beyer, L., Kolesnikov, A., Weissenborn, D., Zhai, X.,
  Unterthiner, T., Dehghani, M., Minderer, M., Heigold, G., Gelly, S., et~al.
\newblock An image is worth 16x16 words: Transformers for image recognition at
  scale.
\newblock \emph{arXiv preprint arXiv:2010.11929}, 2020.

\bibitem[Du et~al.(2020)Du, Chang, Bhunia, Xie, Ma, Song, and Guo]{du2020fine}
Du, R., Chang, D., Bhunia, A.~K., Xie, J., Ma, Z., Song, Y.-Z., and Guo, J.
\newblock Fine-grained visual classification via progressive multi-granularity
  training of jigsaw patches.
\newblock In \emph{European Conference on Computer Vision}, pp.\  153--168,
  2020.

\bibitem[Eilertsen et~al.(2020)Eilertsen, J{\"o}nsson, Ropinski, Unger, and
  Ynnerman]{eilertsen2020classifying}
Eilertsen, G., J{\"o}nsson, D., Ropinski, T., Unger, J., and Ynnerman, A.
\newblock Classifying the classifier: dissecting the weight space of neural
  networks.
\newblock \emph{arXiv preprint arXiv:2002.05688}, 2020.

\bibitem[Fazel(2002)]{fazel2002matrix}
Fazel, M.
\newblock \emph{Matrix rank minimization with applications}.
\newblock PhD thesis, PhD thesis, Stanford University, 2002.

\bibitem[Garg et~al.(2020)Garg, Wu, Balakrishnan, and Lipton]{garg2020unified}
Garg, S., Wu, Y., Balakrishnan, S., and Lipton, Z.
\newblock A unified view of label shift estimation.
\newblock In \emph{Advances in Neural Information Processing Systems}, pp.\
  3290--3300, 2020.

\bibitem[Garg et~al.(2022)Garg, Balakrishnan, Lipton, Neyshabur, and
  Sedghi]{garg2022leveraging}
Garg, S., Balakrishnan, S., Lipton, Z.~C., Neyshabur, B., and Sedghi, H.
\newblock Leveraging unlabeled data to predict out-of-distribution performance.
\newblock In \emph{International Conference on Learning Representations}, 2022.

\bibitem[Gong et~al.(2021)Gong, Lin, Yao, Dietterich, Divakaran, and
  Gervasio]{gong2021confidence}
Gong, Y., Lin, X., Yao, Y., Dietterich, T.~G., Divakaran, A., and Gervasio, M.
\newblock Confidence calibration for domain generalization under covariate
  shift.
\newblock In \emph{Proceedings of the IEEE/CVF International Conference on
  Computer Vision}, pp.\  8958--8967, 2021.

\bibitem[Guillory et~al.(2021)Guillory, Shankar, Ebrahimi, Darrell, and
  Schmidt]{guillory2021predicting}
Guillory, D., Shankar, V., Ebrahimi, S., Darrell, T., and Schmidt, L.
\newblock Predicting with confidence on unseen distributions.
\newblock In \emph{Proceedings of the IEEE/CVF International Conference on
  Computer Vision}, pp.\  1134--1144, 2021.

\bibitem[Guo et~al.(2017)Guo, Pleiss, Sun, and Weinberger]{guo2017calibration}
Guo, C., Pleiss, G., Sun, Y., and Weinberger, K.~Q.
\newblock On calibration of modern neural networks.
\newblock In \emph{Proc. ICML}, pp.\  1321--1330, 2017.

\bibitem[Gupta et~al.(2021)Gupta, Rahimi, Ajanthan, Mensink, Sminchisescu, and
  Hartley]{gupta2020calibration}
Gupta, K., Rahimi, A., Ajanthan, T., Mensink, T., Sminchisescu, C., and
  Hartley, R.
\newblock Calibration of neural networks using splines.
\newblock In \emph{International Conference on Learning Representations}, 2021.

\bibitem[He et~al.(2016)He, Zhang, Ren, and Sun]{he2016deep}
He, K., Zhang, X., Ren, S., and Sun, J.
\newblock Deep residual learning for image recognition.
\newblock In \emph{Proceedings of the IEEE conference on computer vision and
  pattern recognition}, pp.\  770--778, 2016.

\bibitem[Hendrycks \& Dietterich(2019)Hendrycks and
  Dietterich]{hendrycks2019robustness}
Hendrycks, D. and Dietterich, T.
\newblock Benchmarking neural network robustness to common corruptions and
  perturbations.
\newblock In \emph{International Conference on Learning Representations}, 2019.

\bibitem[Hendrycks \& Gimpel(2017)Hendrycks and Gimpel]{hendrycks2017baseline}
Hendrycks, D. and Gimpel, K.
\newblock A baseline for detecting misclassified and out-of-distribution
  examples in neural networks.
\newblock In \emph{International Conference on Learning Representations}, 2017.

\bibitem[Hendrycks et~al.(2021)Hendrycks, Basart, Mu, Kadavath, Wang, Dorundo,
  Desai, Zhu, Parajuli, Guo, et~al.]{hendrycks2021many}
Hendrycks, D., Basart, S., Mu, N., Kadavath, S., Wang, F., Dorundo, E., Desai,
  R., Zhu, T., Parajuli, S., Guo, M., et~al.
\newblock The many faces of robustness: A critical analysis of
  out-of-distribution generalization.
\newblock In \emph{Proceedings of the IEEE/CVF International Conference on
  Computer Vision}, pp.\  8340--8349, 2021.

\bibitem[Jiang et~al.(2019{\natexlab{a}})Jiang, Krishnan, Mobahi, and
  Bengio]{jiang2018predicting}
Jiang, Y., Krishnan, D., Mobahi, H., and Bengio, S.
\newblock Predicting the generalization gap in deep networks with margin
  distributions.
\newblock In \emph{International Conference on Learning Representations},
  2019{\natexlab{a}}.

\bibitem[Jiang et~al.(2019{\natexlab{b}})Jiang, Neyshabur, Mobahi, Krishnan,
  and Bengio]{jiang2019fantastic}
Jiang, Y., Neyshabur, B., Mobahi, H., Krishnan, D., and Bengio, S.
\newblock Fantastic generalization measures and where to find them.
\newblock In \emph{International Conference on Learning Representations},
  2019{\natexlab{b}}.

\bibitem[Karandikar et~al.(2021)Karandikar, Cain, Tran, Lakshminarayanan,
  Shlens, Mozer, and Roelofs]{karandikar2021soft}
Karandikar, A., Cain, N., Tran, D., Lakshminarayanan, B., Shlens, J., Mozer,
  M.~C., and Roelofs, B.
\newblock Soft calibration objectives for neural networks.
\newblock In \emph{Advances in Neural Information Processing Systems}, pp.\
  29768--29779, 2021.

\bibitem[Kendall(1948)]{kendall1948rank}
Kendall, M.~G.
\newblock Rank correlation methods.
\newblock 1948.

\bibitem[Kolesnikov et~al.(2020)Kolesnikov, Beyer, Zhai, Puigcerver, Yung,
  Gelly, and Houlsby]{kolesnikov2020big}
Kolesnikov, A., Beyer, L., Zhai, X., Puigcerver, J., Yung, J., Gelly, S., and
  Houlsby, N.
\newblock Big transfer (bit): General visual representation learning.
\newblock In \emph{European conference on computer vision}, pp.\  491--507,
  2020.

\bibitem[Krause et~al.(2010)Krause, Perona, and
  Gomes]{krause2010discriminative}
Krause, A., Perona, P., and Gomes, R.
\newblock Discriminative clustering by regularized information maximization.
\newblock \emph{Advances in neural information processing systems}, 23, 2010.

\bibitem[Krishnan \& Tickoo(2020)Krishnan and Tickoo]{krishnan2020improving}
Krishnan, R. and Tickoo, O.
\newblock Improving model calibration with accuracy versus uncertainty
  optimization.
\newblock In \emph{Advances in Neural Information Processing Systems},
  volume~33, pp.\  18237--18248, 2020.

\bibitem[Krizhevsky et~al.(2009)Krizhevsky, Hinton,
  et~al.]{krizhevsky2009learning}
Krizhevsky, A., Hinton, G., et~al.
\newblock Learning multiple layers of features from tiny images.
\newblock 2009.

\bibitem[Liang et~al.(2020)Liang, Hu, and Feng]{liang2020we}
Liang, J., Hu, D., and Feng, J.
\newblock Do we really need to access the source data? source hypothesis
  transfer for unsupervised domain adaptation.
\newblock In \emph{International Conference on Machine Learning}, pp.\
  6028--6039. PMLR, 2020.

\bibitem[Lipton et~al.(2018)Lipton, Wang, and Smola]{lipton2018detecting}
Lipton, Z., Wang, Y.-X., and Smola, A.
\newblock Detecting and correcting for label shift with black box predictors.
\newblock In \emph{International conference on machine learning}, pp.\
  3122--3130, 2018.

\bibitem[Liu et~al.(2021)Liu, Lin, Cao, Hu, Wei, Zhang, Lin, and
  Guo]{liu2021swin}
Liu, Z., Lin, Y., Cao, Y., Hu, H., Wei, Y., Zhang, Z., Lin, S., and Guo, B.
\newblock Swin transformer: Hierarchical vision transformer using shifted
  windows.
\newblock In \emph{Proceedings of the IEEE/CVF International Conference on
  Computer Vision}, pp.\  10012--10022, 2021.

\bibitem[Liu et~al.(2022)Liu, Mao, Wu, Feichtenhofer, Darrell, and
  Xie]{liu2022convnet}
Liu, Z., Mao, H., Wu, C.-Y., Feichtenhofer, C., Darrell, T., and Xie, S.
\newblock A convnet for the 2020s.
\newblock In \emph{Proceedings of the IEEE Conference on Computer Vision and
  Pattern Recognition}, 2022.

\bibitem[Madani et~al.(2004)Madani, Pennock, and Flake]{madani2004co}
Madani, O., Pennock, D., and Flake, G.
\newblock Co-validation: Using model disagreement on unlabeled data to validate
  classification algorithms.
\newblock In \emph{Advances in neural information processing systems}, pp.\
  873--880, 2004.

\bibitem[Miller et~al.(2021)Miller, Taori, Raghunathan, Sagawa, Koh, Shankar,
  Liang, Carmon, and Schmidt]{miller2021accuracy}
Miller, J.~P., Taori, R., Raghunathan, A., Sagawa, S., Koh, P.~W., Shankar, V.,
  Liang, P., Carmon, Y., and Schmidt, L.
\newblock Accuracy on the line: on the strong correlation between
  out-of-distribution and in-distribution generalization.
\newblock In \emph{International Conference on Machine Learning}, pp.\
  7721--7735, 2021.

\bibitem[Minderer et~al.(2021)Minderer, Djolonga, Romijnders, Hubis, Zhai,
  Houlsby, Tran, and Lucic]{minderer2021revisiting}
Minderer, M., Djolonga, J., Romijnders, R., Hubis, F., Zhai, X., Houlsby, N.,
  Tran, D., and Lucic, M.
\newblock Revisiting the calibration of modern neural networks.
\newblock In \emph{Advances in Neural Information Processing Systems}, pp.\
  15682--15694, 2021.

\bibitem[Mintun et~al.(2021)Mintun, Kirillov, and Xie]{mintun2021interaction}
Mintun, E., Kirillov, A., and Xie, S.
\newblock On interaction between augmentations and corruptions in natural
  corruption robustness.
\newblock In \emph{Advances in Neural Information Processing Systems}, 2021.

\bibitem[Nagelkerke et~al.(1991)]{nagelkerke1991note}
Nagelkerke, N.~J. et~al.
\newblock A note on a general definition of the coefficient of determination.
\newblock \emph{Biometrika}, 78\penalty0 (3):\penalty0 691--692, 1991.

\bibitem[Neyshabur et~al.(2017)Neyshabur, Bhojanapalli, McAllester, and
  Srebro]{neyshabur2017exploring}
Neyshabur, B., Bhojanapalli, S., McAllester, D., and Srebro, N.
\newblock Exploring generalization in deep learning.
\newblock In \emph{Advances in neural information processing systems}, pp.\
  5947--5956, 2017.

\bibitem[Ovadia et~al.(2019)Ovadia, Fertig, Ren, Nado, Sculley, Nowozin,
  Dillon, Lakshminarayanan, and Snoek]{ovadia2019can}
Ovadia, Y., Fertig, E., Ren, J., Nado, Z., Sculley, D., Nowozin, S., Dillon,
  J., Lakshminarayanan, B., and Snoek, J.
\newblock Can you trust your model's uncertainty? evaluating predictive
  uncertainty under dataset shift.
\newblock In \emph{Advances in Neural Information Processing Systems}, 2019.

\bibitem[Platanios et~al.(2017)Platanios, Poon, Mitchell, and
  Horvitz]{platanios2017estimating}
Platanios, E., Poon, H., Mitchell, T.~M., and Horvitz, E.~J.
\newblock Estimating accuracy from unlabeled data: A probabilistic logic
  approach.
\newblock In \emph{Advances in Neural Information Processing Systems}, pp.\
  4361--4370, 2017.

\bibitem[Platanios et~al.(2016)Platanios, Dubey, and
  Mitchell]{platanios2016estimating}
Platanios, E.~A., Dubey, A., and Mitchell, T.
\newblock Estimating accuracy from unlabeled data: A bayesian approach.
\newblock In \emph{International Conference on Machine Learning}, pp.\
  1416--1425, 2016.

\bibitem[Recht et~al.(2010)Recht, Fazel, and Parrilo]{recht2010guaranteed}
Recht, B., Fazel, M., and Parrilo, P.~A.
\newblock Guaranteed minimum-rank solutions of linear matrix equations via
  nuclear norm minimization.
\newblock \emph{SIAM review}, 52\penalty0 (3):\penalty0 471--501, 2010.

\bibitem[Recht et~al.(2018)Recht, Roelofs, Schmidt, and
  Shankar]{recht2018cifar}
Recht, B., Roelofs, R., Schmidt, L., and Shankar, V.
\newblock Do cifar-10 classifiers generalize to cifar-10?
\newblock \emph{arXiv preprint arXiv:1806.00451}, 2018.

\bibitem[Recht et~al.(2019)Recht, Roelofs, Schmidt, and
  Shankar]{recht2019imagenet}
Recht, B., Roelofs, R., Schmidt, L., and Shankar, V.
\newblock Do imagenet classifiers generalize to imagenet?
\newblock In \emph{International Conference on Machine Learning}, pp.\
  5389--5400. PMLR, 2019.

\bibitem[Schiff et~al.(2021)Schiff, Quanz, Das, and Chen]{schiff2021predicting}
Schiff, Y., Quanz, B., Das, P., and Chen, P.-Y.
\newblock Predicting deep neural network generalization with perturbation
  response curves.
\newblock In \emph{Advances in Neural Information Processing Systems}, 2021.

\bibitem[Shi \& Sha(2012)Shi and Sha]{shi2012information}
Shi, Y. and Sha, F.
\newblock Information-theoretical learning of discriminative clusters for
  unsupervised domain adaptation.
\newblock \emph{arXiv preprint arXiv:1206.6438}, 2012.

\bibitem[Simonyan \& Zisserman(2014)Simonyan and Zisserman]{simonyan2014very}
Simonyan, K. and Zisserman, A.
\newblock Very deep convolutional networks for large-scale image recognition.
\newblock \emph{arXiv preprint arXiv:1409.1556}, 2014.

\bibitem[Sugiyama \& Kawanabe(2012)Sugiyama and Kawanabe]{sugiyama2012machine}
Sugiyama, M. and Kawanabe, M.
\newblock \emph{Machine learning in non-stationary environments: Introduction
  to covariate shift adaptation}.
\newblock MIT press, 2012.

\bibitem[Sun et~al.(2022)Sun, Lu, and Ling]{sun2022prior}
Sun, T., Lu, C., and Ling, H.
\newblock Prior knowledge guided unsupervised domain adaptation.
\newblock In \emph{European Conference on Computer Vision}, 2022.

\bibitem[Tang et~al.(2020)Tang, Chen, and Jia]{tang2020unsupervised}
Tang, H., Chen, K., and Jia, K.
\newblock Unsupervised domain adaptation via structurally regularized deep
  clustering.
\newblock In \emph{Proceedings of the IEEE/CVF conference on computer vision
  and pattern recognition}, pp.\  8725--8735, 2020.

\bibitem[Taori et~al.(2020)Taori, Dave, Shankar, Carlini, Recht, and
  Schmidt]{taori2020measuring}
Taori, R., Dave, A., Shankar, V., Carlini, N., Recht, B., and Schmidt, L.
\newblock Measuring robustness to natural distribution shifts in image
  classification.
\newblock \emph{Advances in Neural Information Processing Systems},
  33:\penalty0 18583--18599, 2020.

\bibitem[Tian et~al.(2020)Tian, Liu, Glaser, Hsu, and Kira]{tian2020posterior}
Tian, J., Liu, Y.-C., Glaser, N., Hsu, Y.-C., and Kira, Z.
\newblock Posterior re-calibration for imbalanced datasets.
\newblock \emph{Advances in Neural Information Processing Systems},
  33:\penalty0 8101--8113, 2020.

\bibitem[Unterthiner et~al.(2020)Unterthiner, Keysers, Gelly, Bousquet, and
  Tolstikhin]{unterthiner2020predicting}
Unterthiner, T., Keysers, D., Gelly, S., Bousquet, O., and Tolstikhin, I.
\newblock Predicting neural network accuracy from weights.
\newblock In \emph{International Conference on Learning Representations}, 2020.

\bibitem[Wah et~al.(2011)Wah, Branson, Welinder, Perona, and
  Belongie]{wah2011caltech}
Wah, C., Branson, S., Welinder, P., Perona, P., and Belongie, S.
\newblock The caltech-ucsd birds-200-2011 dataset.
\newblock 2011.

\bibitem[Wang et~al.(2019)Wang, Ge, Lipton, and Xing]{wang2019learning}
Wang, H., Ge, S., Lipton, Z., and Xing, E.~P.
\newblock Learning robust global representations by penalizing local predictive
  power.
\newblock In \emph{Advances in Neural Information Processing Systems}, pp.\
  10506--10518, 2019.

\bibitem[Wang et~al.(2020)Wang, Chen, Wang, Long, and
  Wang]{wang2020progressive}
Wang, S., Chen, X., Wang, Y., Long, M., and Wang, J.
\newblock Progressive adversarial networks for fine-grained domain adaptation.
\newblock In \emph{Proceedings of the IEEE/CVF conference on computer vision
  and pattern recognition}, pp.\  9213--9222, 2020.

\bibitem[Wei et~al.(2021)Wei, Song, Mac~Aodha, Wu, Peng, Tang, Yang, and
  Belongie]{wei2021fine}
Wei, X.-S., Song, Y.-Z., Mac~Aodha, O., Wu, J., Peng, Y., Tang, J., Yang, J.,
  and Belongie, S.
\newblock Fine-grained image analysis with deep learning: A survey.
\newblock \emph{IEEE Transactions on Pattern Analysis and Machine
  Intelligence}, 2021.

\bibitem[Wightman(2019)]{rw2019timm}
Wightman, R.
\newblock Pytorch image models.
\newblock \url{https://github.com/rwightman/pytorch-image-models}, 2019.

\bibitem[Yang et~al.(2021)Yang, van~de Weijer, Herranz, Jui,
  et~al.]{yang2021exploiting}
Yang, S., van~de Weijer, J., Herranz, L., Jui, S., et~al.
\newblock Exploiting the intrinsic neighborhood structure for source-free
  domain adaptation.
\newblock In \emph{Advances in Neural Information Processing Systems}, pp.\
  29393--29405, 2021.

\bibitem[Yang et~al.(2022)Yang, Wang, Wang, Jui, and van~de
  Weijer]{yang2022attracting}
Yang, S., Wang, Y., Wang, K., Jui, S., and van~de Weijer, J.
\newblock Attracting and dispersing: A simple approach for source-free domain
  adaptation.
\newblock In \emph{Advances in Neural Information Processing Systems}, 2022.

\bibitem[Yu et~al.(2022)Yu, Yang, Wei, Ma, and Steinhardt]{yu2022predicting}
Yu, Y., Yang, Z., Wei, A., Ma, Y., and Steinhardt, J.
\newblock Predicting out-of-distribution error with the projection norm.
\newblock In \emph{Advances in Neural Information Processing Systems}, 2022.

\end{thebibliography}
\bibliographystyle{icml2023}

\appendix
\onecolumn

\section{Nuclear Norm}\label{supp:nuclear}
Let $\mP\in\mathbb{R}^{n_t \times k}$ denote the prediction matrix of $f$ on $\mathcal{D}_\mathrm{u}^{T}$, nuclear norm $||\mP||_*$ is the sum of singular values of $\mP$. 
Nuclear norm is the tightest convex envelope of rank function within the unit ball \citep{fazel2002matrix}. 
{A larger nuclear norm implies more classes are predicted and involved, indicating higher prediction dispersity.}
In addition, nuclear norm $||\mP||_*$ and Frobenius norm~$||\mP||_F=\sqrt{Trace(\mP^\intercal\mP})$ can bound each other \citep{recht2010guaranteed,fazel2002matrix}.
More specifically, they have the following relationship:~$1/\sqrt{d}||\mP||_* \leq ||\mP||_F \leq ||\mP||_* \leq \sqrt{d}||\mP||_F$,~where $d=min(n_t, k)$. In our work, because $\mP$ consists of softmax vectors, its Frobenius norm is bound by $||\mP||_F \leq \sqrt{n_t}$.

Frobenius norm $||\mP||_F$ reflects prediction confidence \cite{cui2020towards}. 
Based on the above relationship, a larger nuclear norm $||\mP||_*$ implies a larger Frobenius norm $||\mP||_F$, indicating a higher prediction confidence.
Therefore, nuclear norm $||\mP||_*$ can be used to characterize both confidence and dispersity of $\mP$. Moreover, nuclear norm $||\mP||_*$ is related to the shape of $\mP$, so we normalized it by its upper bound~$\sqrt{d \cdot n_t}$ and obtain $\widehat{||\mP||_*} = ||\mP||_*/\sqrt{d \cdot n_t}$. In our work, we use $\widehat{||\mP||_*}$ to measure the confidence and dispersity of the prediction matrix.

\section{Difference From Domain Adaptation}
Unsupervised accuracy estimation and unsupervised domain adaptation are significantly different tasks.
\textbf{First}, the two tasks have different settings and goals. Unsupervised domain adaptation considers a fixed pair of source-target datasets. Given labeled source data and unlabeled target data, its goal is to learn an adaptive model that generalizes well to the unlabeled target domain. In comparison, unsupervised accuracy estimation considers various target datasets and a trained model. This goal is not to adapt the model to the target data. Instead, its goal is to estimate the performance of the trained and fixed model on various unlabeled test sets.
\textbf{Second}, the two tasks have different research directions. Unsupervised domain adaptation works develop domain adaptive algorithms to eliminate domain discrepancy. In contrast, unsupervised accuracy estimation methods typically derive model-based distribution statistics of test sets (\textit{e.g.}, DoC and ATC) for the accuracy estimation.

\section{Experimental Setup} \label{sec:models}

\subsection{Models}
\textbf{ImageNet.} 
Models are provided by PyTorch Image Models (timm-1.5) \cite{rw2019timm}. They are either trained or fine-tuned on the ImageNet-1k training set \cite{deng2009imagenet}.

\textbf{CIFAR-10.} We train models using the implementations from \textcolor{blue}{https://github.com/chenyaofo/pytorch-cifar-models}. CIFAR-$\bar{C}$-Rand is generated with the $10$ new corruptions of ImageNet-$\bar{C}$ \citep{mintun2021interaction} that are \textit{perceptually dissimilar} to ImageNet-C.
We apply random corruptions following \textcolor{blue}{https://github.com/facebookresearch/augmentation-corruption}.

\textbf{CUB-200.} We train CIFAR models using the implementations from \textcolor{blue}{https://github.com/PRIS-CV/PMG-Progressive-Multi-Granularity-Training}. CUB-200-C is generated based on the implementations from \textcolor{blue}{https://github.com/hendrycks/robustness}.

\subsection{Datasets} 
The datasets we use are standard benchmarks, which are publicly available. We have double-checked their license. We list their open-source as follows.

\textbf{CIFAR-10} \cite{krizhevsky2009learning} (\textcolor{blue}{https://www.cs.toronto.edu/~kriz/cifar.html});\\
\textbf{CIFAR-10-C} \cite{hendrycks2019robustness} (\textcolor{blue}{https://github.com/hendrycks/robustness}); \\
\textbf{CIFAR-10.1} \cite{recht2018cifar} (\textcolor{blue}{https://github.com/modestyachts/CIFAR-10.1});\\
\textbf{CINIC} \cite{chrabaszcz2017downsampled} (\textcolor{blue}{https://github.com/BayesWatch/cinic-10}).

\textbf{ImageNet-Validation} \cite{deng2009imagenet} (\textcolor{blue}{https://www.image-net.org}); \\
\textbf{ImageNet-V2-A/B/C} \cite{recht2019imagenet} (\textcolor{blue}{https://github.com/modestyachts/ImageNetV2}); \\
\textbf{ImageNet-Corruption} \cite{hendrycks2019robustness} (\textcolor{blue}{https://github.com/hendrycks/robustness});\\
\textbf{ImageNet-Sketch} \cite{wang2019learning} (\textcolor{blue}{https://github.com/HaohanWang/ImageNet-Sketch});\\
\textbf{ImageNet-Rendition} \cite{hendrycks2021many} (\textcolor{blue}{https://github.com/hendrycks/imagenet-r});\\
\textbf{ObjectNet} \cite{barbu2019objectnet} (\textcolor{blue}{https://objectnet.dev}).

\textbf{CUB-200-2011} \cite{wah2011caltech} (\textcolor{blue}{https://www.vision.caltech.edu/datasets/cub\_200\_2011}).
\\
\textbf{CUB-Paintings} \cite{wang2020progressive} (\textcolor{blue}{https://github.com/thuml/PAN}).

\subsection{Computation Resources} 
We run all experiment on one 3090Ti with PyTorch (1.11.0+cu113). CPU is \textcolor{black}{AMD Ryzen 9 5900X 12-Core Processor}. 

\subsection{Experimental Detail}

\begin{figure}[!ht]
    \begin{center}
    \includegraphics[width=0.85\linewidth]{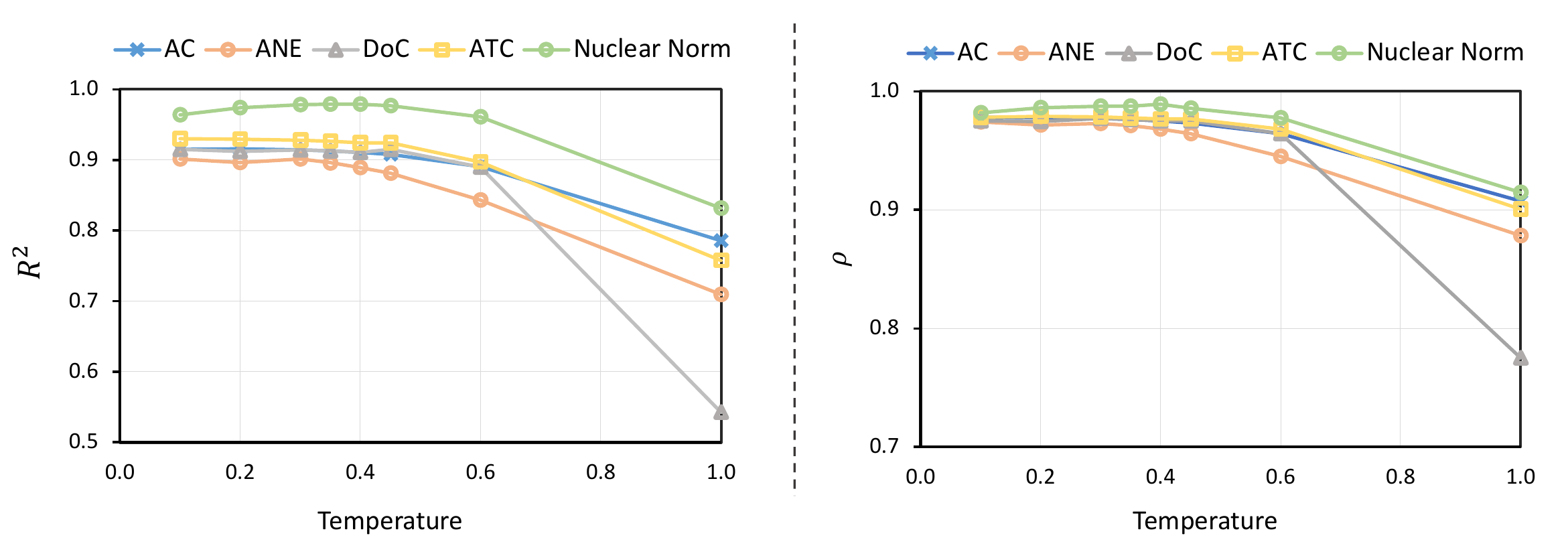}
        \vspace{-0.5cm}
    \caption{\textbf{Effect of temperature for all methods.} We report the correlation results (both $R^2$ and $\rho$) using various temperature of softmax. 
    We show that a small temperature ($0.2$ to $0.45$) helps for all methods. Moreover, when using different temperature values, nuclear norm consistently exhibits a stronger correlation than other methods.
    }\label{fig:temp}
    \vspace{-0.6cm}
    \end{center}
\end{figure}

\textbf{(I) Effect of temperature. }
We empirically find that using a small temperature for softmax is helpful for all methods. Therefore, we use temperature of $0.4$ for all methods in the experiment. 
We show the effect of temperature in term of correlation strength ($R^2$ and $\rho$) in Fig. \ref{fig:temp}. 
We have two observations. \textbf{First}, using a small temperature (\emph{e.g.}, $0.4$) helps for all methods including nuclear norm, ATC and DoC. The correlation results are stable when temperature ranges from $0.2$ to $0.45$.
\textbf{Second}, when using various temperature values, nuclear norm consistently achieve stronger correlation.

\begin{figure}[!ht]
    \begin{center}
    \includegraphics[width=1\linewidth]{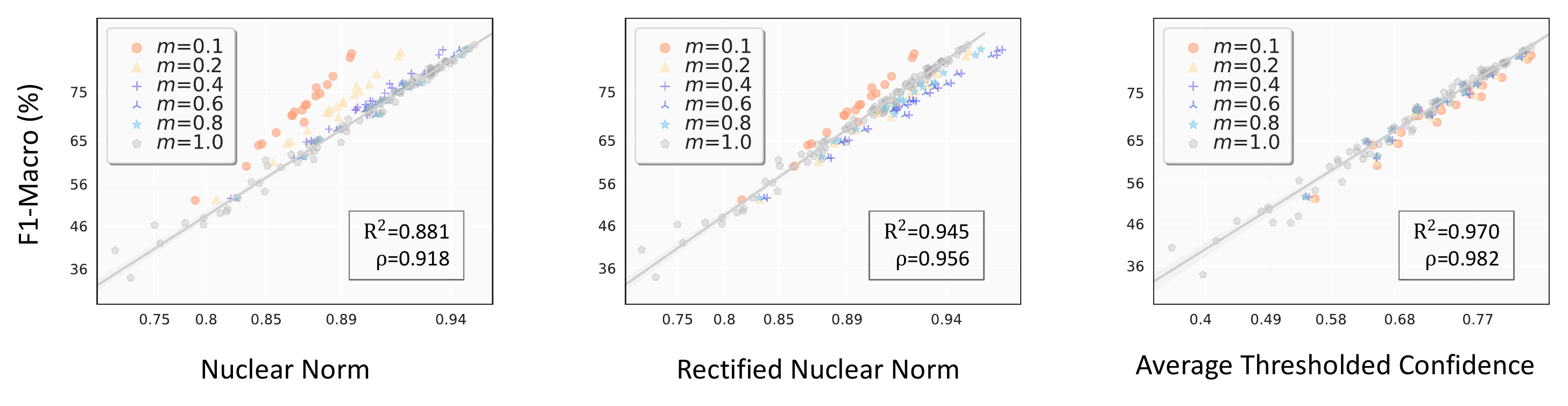}
        \vspace{-0.6cm}
    \caption{\wj{\textbf{Effect of rectified nuclear norm.} Under imbalanced test sets, we relax the regularization of nuclear norm on ``tail" classes (rectified nuclear norm).
    We conduct a correlation study on \textit{imbalanced} ImageNet-C using \textbf{ViT}.
    We observe that rectified nuclear norm can improve the nuclear norm under imbalanced test sets.
    }}\label{fig:r_nm}
     \vspace{-12pt}
    \end{center}
\end{figure}

\begin{wraptable}{r}{0.46\textwidth}
\small
    \centering
    \setlength{\tabcolsep}{2pt}
    \begin{tabular}{c c c c}
        \toprule
         \textbf{Correlation} &  \multicolumn{1}{c}{\textbf{ProjNorm}} & \multicolumn{1}{c}{\textbf{ALine-D}} & \multicolumn{1}{c}{\textbf{Nuclear Norm}}\\ 
        \midrule
         $\rho$ & 0.980 & 0.995 & \textbf{0.997}  \\
         $R^2$ & 0.973 & 0.974 & \textbf{0.990}  \\
        \bottomrule
    \end{tabular}
    \caption{\textbf{Method comparison under CIFAR-10 setup.} We report the average correlation strength (Spearman's rank correlation $\rho$ and coefficients of determination $R^2$).
    }
    \label{tab:proj}
\end{wraptable}
\textbf{(III) Comparison with ALine-D \citep{baek2022agreement} and ProjNorm \citep{yu2022predicting}}. \wj{For a fair comparison, we follow the same setting as \citep{baek2022agreement} and report the results using ResNet18 on CIFAR-10-C. As shown in Table \ref{tab:proj}, we observe that nuclear norm gains stronger correlation strength than the two methods. It achieves $0.997$ and $0.990$ in rank correlation ($\rho$) and coefficients of determination ($R^2$), respectively. Furthermore, we would like to mention that ALine-D \citep{baek2022agreement} requires a set of models for accuracy estimation. ProjNorm \citep{yu2022predicting} requires fine-tuning a pre-trained network on each OOD test set with pseudo-labels. In contrast, nuclear Norm is more efficient: it is computed on a classifier's prediction matrix on each unlabeled test set.}

\wj{\textbf{(II) Rectified nuclear norm for imbalanced test sets}. We tried to relax the regularization of nuclear norm under imbalanced test sets. Nuclear Norm encourages the predictions to be well-distributed across all classes. 
For imbalanced test sets, we can relax this regularization on the tail classes. That is, we mainly consider the prediction dispersity of head classes. 

To achieve this, we explored one intuitive way to rectify the nuclear norm: we modify the normalization (i.e., upper bound) of the nuclear norm. Specifically, we revise the normalization from $\sqrt{\min(n_t, k) * n_t}$ to $\sqrt{\min(n_t, k_\text{head}) * n_t}$, where $k_\text{head}$ is the number of major classes regularised by nuclear norm. We conducted experiment under ImageNet setup (k=1000) and empirically set $k_\text{head}$ based on the imbalanced intensity $r_m$ (the ratio between the number of last 10 ``tail" classes and the number of top 10 ``head" classes):  $k_\text{head} = k - (1 - r_m) *80$. To estimate imbalanced intensity, we use BBSE \citep{lipton2018detecting} to estimate the class distribution.

In Figure \ref{fig:r_nm}, we show that our attempt (rectified nuclear norm) can improve nuclear norm. We would like to view the above experiment as a starting point that inspires more research on the rectification of nuclear norm for strong imbalanced test sets.}

\wj{\textbf{(IV) Additional observations.} \textbf{First}, ObjectNet of ImageNet setup is built in a bias-controlled manner (with controls for rotation, background, and viewpoint). We observe that its images are often confidently misclassified, which makes predictions with the high nuclear norm. We believe this is why ObjectNet is always off the linear line.
\textbf{Second}, for all accuracy estimation methods, they can well capture the model performance is high (top-right region of each figure). However, when model accuracy is low (bottom-left), existing methods cannot make reasonable estimations, especially under CIFAR-10 and CUB-200. In contrast, nuclear norm can well handle the low-accuracy region by additionally considering the prediction dispersity. To improve the accuracy estimation, it would be helpful to further consider the characteristics of predictions when the model performs poorly.
\textbf{Third}, in Figures \ref{fig:img}, \ref{fig:cifar}, and \ref{fig:cub}, we observe that the real-world test sets (\textit{e.g.}, ImageNet-R, CINIC, and CUB-P) scatter around the linear lines fit on synthetic datasets. This indicates that both real-world and synthetic datasets follow a similar linear trend. This gives an interesting hint: we can use synthetic datasets to simulate and capture the distributions of real-world test sets.}

\subsection{More Correlation Results}

\subsubsection{ImageNet Setup}

\begin{figure}[!ht]
    \begin{center}
    \includegraphics[width=0.9\linewidth]{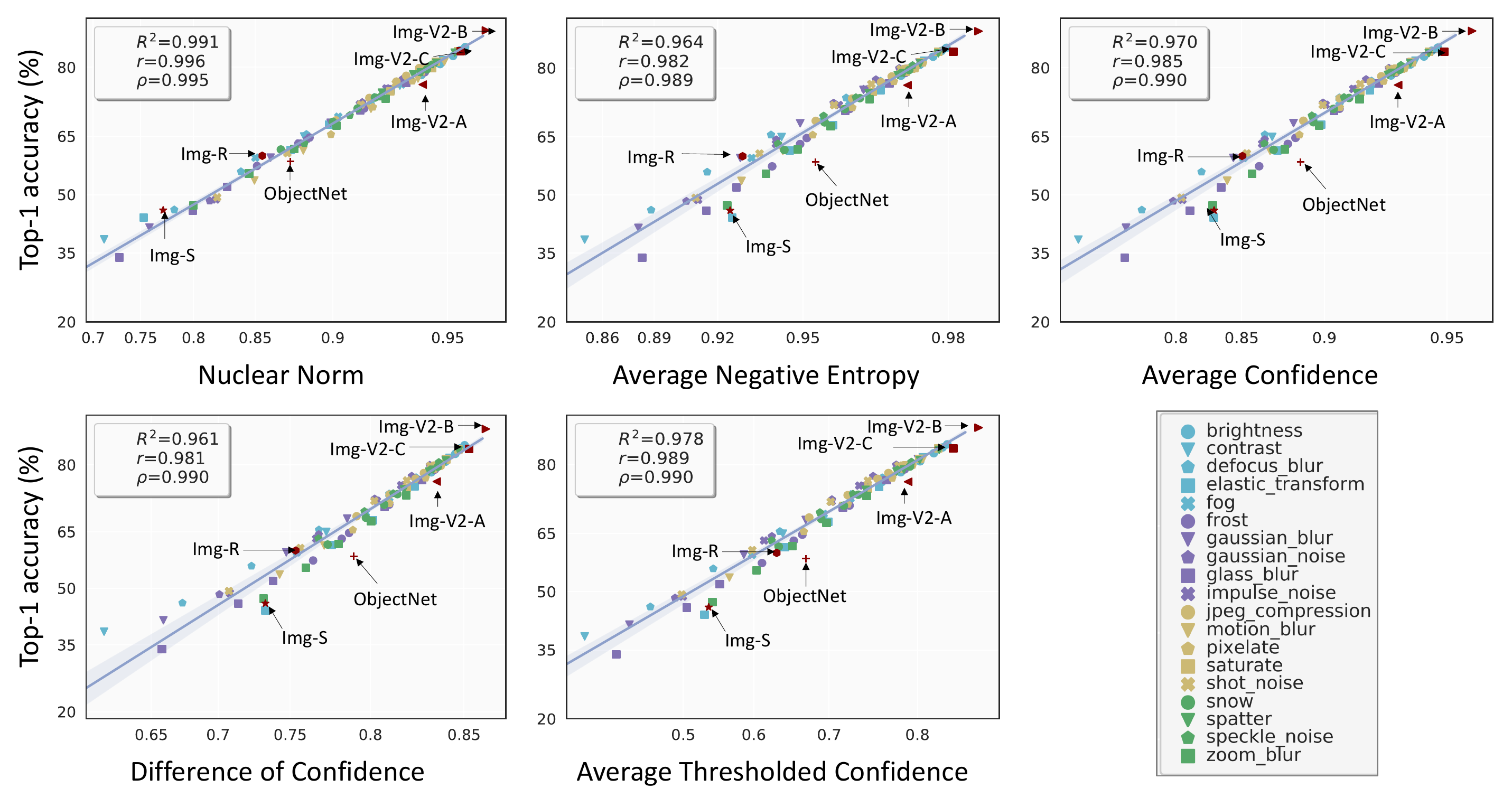}
    \caption{\textbf{Correlation study under the ImageNet setup.} We plot the actual accuracy of \textbf{\textit{ViT}} and \textcolor{black}{five measures including nuclear norm and four competing methods}. %
    Different shapes in each sub-figure represents different test sets. The straight lines are calculated by linear regression fit on synthetic datasets of ImageNet-C. 
    We list the $19$ types of corruptions in ImageNet-C using different shapes and colors in the bottom right figure.
    We also mark the $6$ real-world datasets in each sub-figure with arrows.
     Compared with other methods, nuclear norm exhibits stronger correlation with accuracy.
     Moreover, with nuclear norm, real-world test sets are closely around the linearly fit line.
    }\label{fig:vit}
    \end{center}
\end{figure}

\begin{figure}[!ht]
    \begin{center}
    \includegraphics[width=1\linewidth]{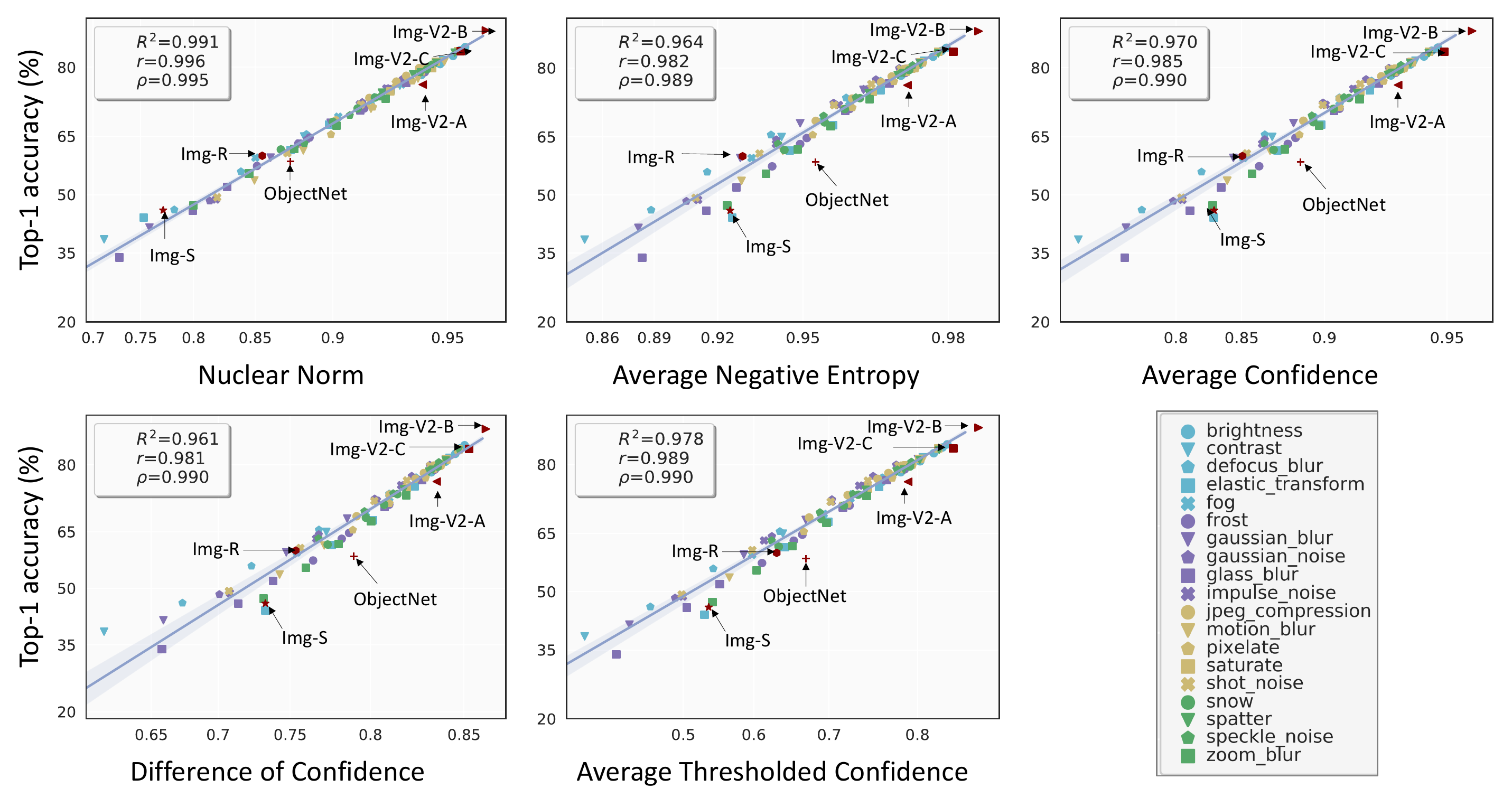}
    \caption{\textbf{Correlation study under the ImageNet setup.} We plot the actual accuracy of \textbf{\textit{BeiT}} and \textcolor{black}{five measures including nuclear norm and four competing methods}.
    }\label{fig:vit}
    \end{center}
    \vspace{-2cm}
\end{figure}

\begin{figure}[!ht]
    \begin{center}
    \includegraphics[width=1\linewidth]{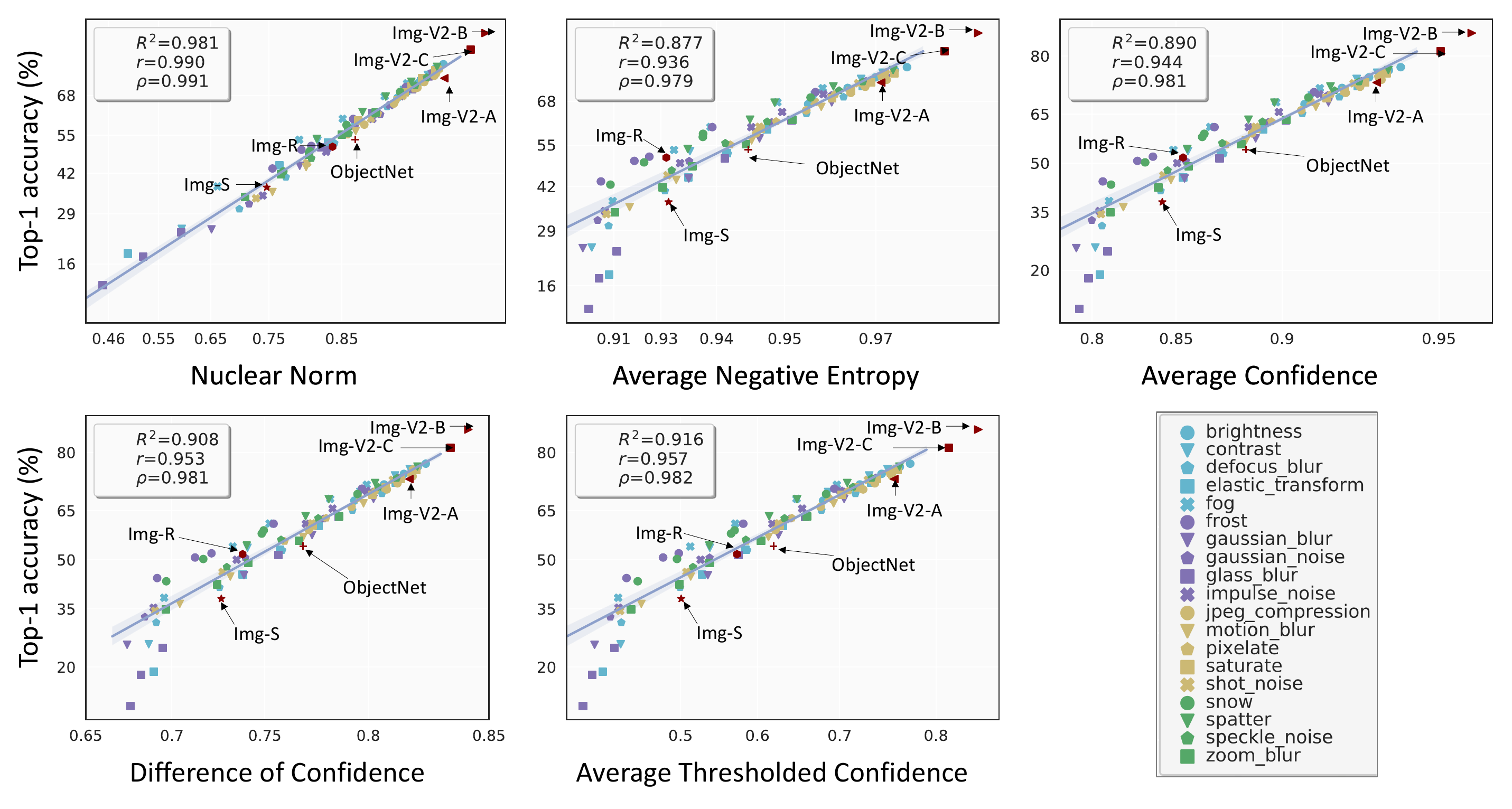}
    \caption{\textbf{Correlation study under the ImageNet setup.} We plot the actual accuracy of \textbf{\textit{Res152-BiT}} and \textcolor{black}{five measures including nuclear norm and four competing methods}.
    }\label{fig:res152}
    \end{center}
\end{figure}

\newpage
\subsubsection{CIFAR-10 and CUB-200 Setups}

\begin{figure}[!ht]
    \begin{center}
    \includegraphics[width=1\linewidth]{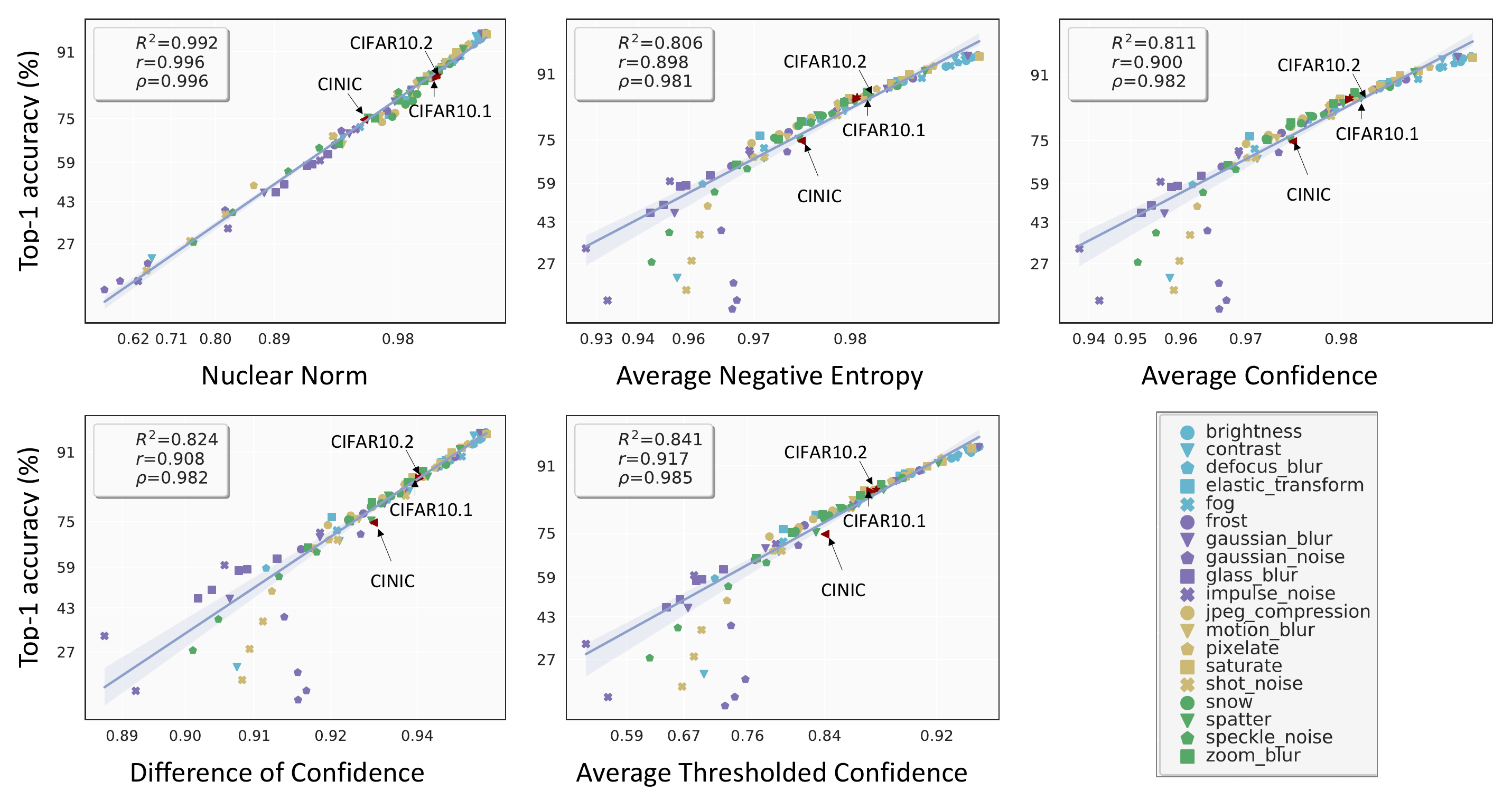}
    \caption{\textbf{Correlation study under the CIFAR-10 setup.} We plot the actual accuracy of \textbf{\textit{RepVGG-A0}} and \textcolor{black}{five measures including nuclear norm and four competing methods}.
     The straight lines are calculated by linear regression fit on synthetic datasets of CIFAR-10-C. 
    We list the $19$ types of corruptions in CIFAR-10-C using different shapes and colors in the bottom right figure.
    We also mark the $3$ real-world datasets in each sub-figure with arrows.
     Compared with other methods, nuclear norm exhibits stronger correlation with accuracy.
    }\label{fig:reg}
    \end{center}
    \vspace{-1cm}
\end{figure}

\begin{figure}[!ht]
    \begin{center}
    \includegraphics[width=1\linewidth]{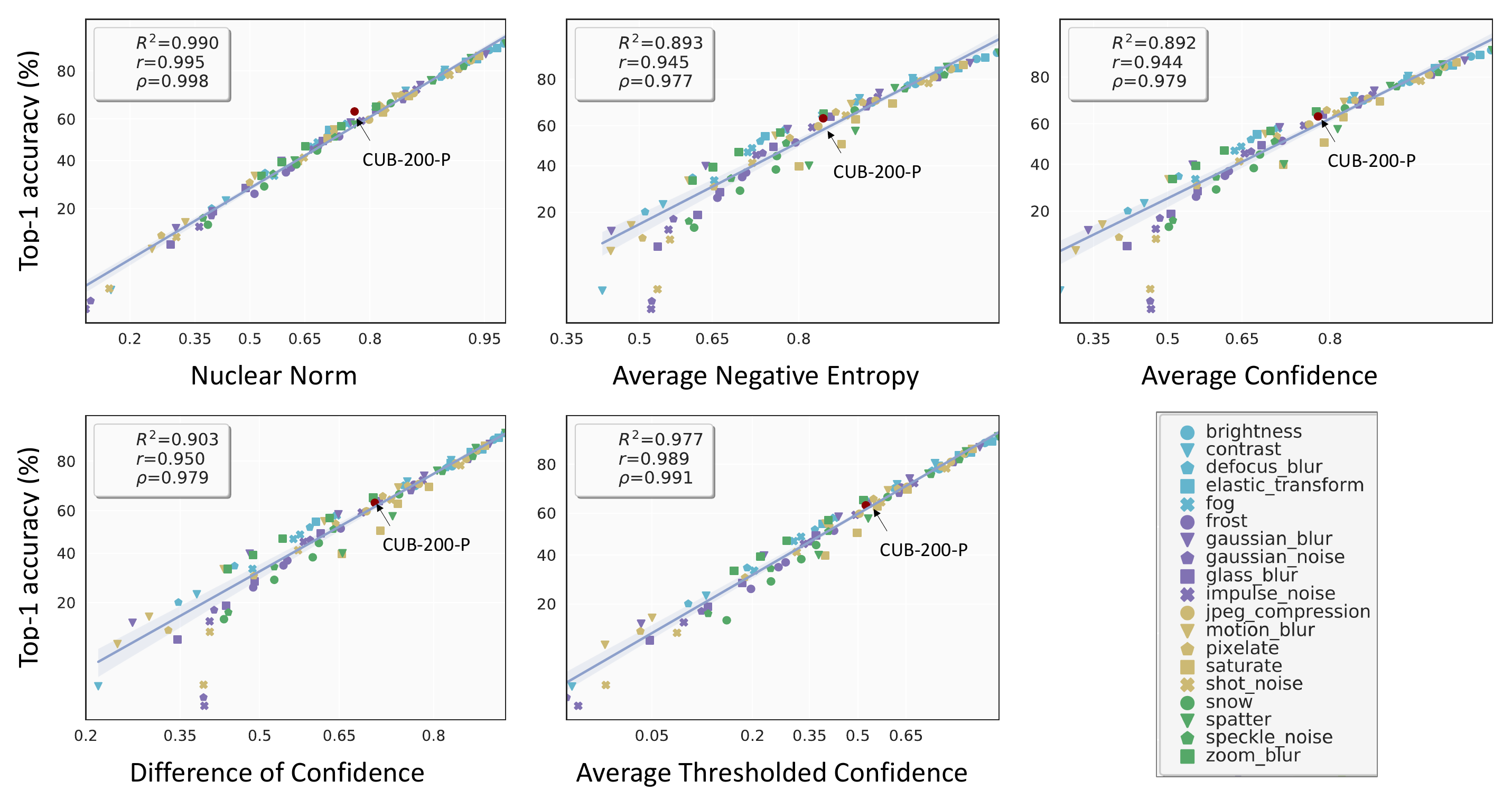}
 
    \caption{\textbf{Correlation study under the CUB-200 setup.} We plot the actual accuracy of \textbf{\textit{PMG}} and \textcolor{black}{five measures including nuclear norm and four competing methods}.
   The straight lines are calculated by linear regression fit on synthetic datasets of CUB-200-C. 
    We list the $19$ types of corruptions in CUB-200-C using different shapes and colors in the bottom right figure.
    We also mark the real-world CUB-200-P in each sub-figure with arrows.
     Compared with other methods, nuclear norm exhibits a stronger correlation with accuracy.
    }\label{fig:pmg}
    \end{center}
\end{figure}

\end{document}